\documentclass[runningheads]{llncs}

\usepackage{eccv}

\usepackage{multirow} 

\usepackage{eccvabbrv}

\usepackage{graphicx}
\usepackage{booktabs}

\usepackage[accsupp]{axessibility}  

\makeatletter
\def\thanks#1{\protected@xdef\@thanks{\@thanks
        \protect\footnotetext{#1}}}
\makeatother

\usepackage{hyperref}

\usepackage{orcidlink}
\usepackage{multirow}
\usepackage{arydshln}
\usepackage{pifont}
\usepackage{threeparttable}
\usepackage{marvosym}

\begin{document}

\title{Ref-AVS: Refer and Segment Objects in Audio-Visual Scenes} 

\titlerunning{Ref-AVS}


\author{Yaoting Wang\inst{1}\inst{\dagger}\orcidlink{0009-0004-5724-5698}\thanks{$^\dagger$ Equal contribution.} \and
Peiwen Sun\inst{2}\inst{\dagger}\orcidlink{0009-0005-3016-8554} \and
Dongzhan Zhou\inst{3}\inst{\dagger}\orcidlink{0000-0001-6568-5440} \and
Guangyao Li\inst{1}\orcidlink{0000-0002-2179-8555} \and
Honggang Zhang\inst{2}\orcidlink{0000-0001-8287-6783}\and
Di Hu\textsuperscript{\Letter}\inst{1,4}\orcidlink{0000-0002-7118-6733} 
\thanks{\textsuperscript{\Letter} Corresponding author.}
}

\authorrunning{Y.~Wang et al.}

\institute{
\textsuperscript{1} Gaoling School of Artificial Intelligence, Renmin University of China, China \\
\email{yaoting.wang@outlook.com} \\
\email{\{guangyaoli,dihu\}@ruc.edu.cn} \\
\textsuperscript{2} Beijing University of Posts and Telecommunications, Beijing, China \\
\email{\{sunpeiwen,zhhg\}@bupt.edu.cn}\\
\textsuperscript{3} Shanghai Artificial Intelligence Laboratory, Shanghai, China \\
\email{zhoudongzhan@pjlab.org.cn}\\
\textsuperscript{4} Engineering Research Center of Next-Generation Search and Recommendation\\
}

\maketitle

\begin{abstract}
  Traditional reference segmentation tasks have predominantly focused on silent visual scenes, neglecting the integral role of multimodal perception and interaction in human experiences. 
  In this work, we introduce a novel task called Reference Audio-Visual Segmentation (Ref-AVS), which seeks to segment objects within the visual domain based on expressions containing multimodal cues. 
  Such expressions are articulated in natural language forms but are enriched with multimodal cues, including audio and visual descriptions.
  To facilitate this research, we construct the first Ref-AVS benchmark, which provides pixel-level annotations for objects described in corresponding multimodal-cue expressions. 
  To tackle the Ref-AVS task, we propose a new method that adequately utilizes multimodal cues to offer precise segmentation guidance.
  Finally, we conduct quantitative and qualitative experiments on three test subsets to compare our approach with existing methods from related tasks. 
  The results demonstrate the effectiveness of our method, highlighting its capability to precisely segment objects using multimodal-cue expressions. Dataset is available at \href{https://gewu-lab.github.io/Ref-AVS}{https://gewu-lab.github.io/Ref-AVS}.
\end{abstract}

\begin{figure*}[h]
  \centering
   \includegraphics[width=1\linewidth]{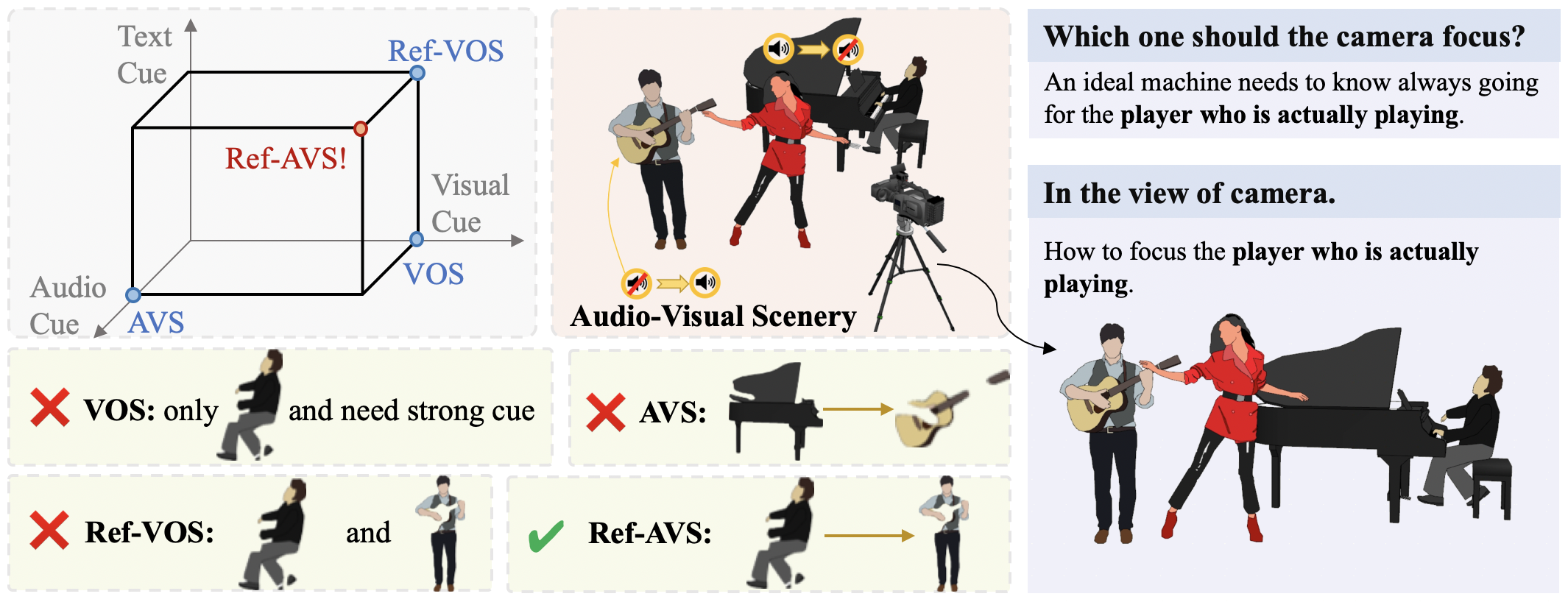}
 
   \caption{Comparison of the Ref-AVS task with other related tasks. Ref-AVS challenges machines to locate objects of interest in the visual space using multimodal cues, just like humans do in the real world.}
 
   \label{fig:teaser}
\end{figure*}
\section{Introduction}
In the real world, visual scenes are often accompanied by diverse \textit{multimodal information}\footnote{To clarify, we refer to the sound and vision in the scene as multimodal information, while the attribute description of audio and vision (\eg, louder and on the right) in the reference expression is named multimodal cues.}, including audio and text modalities. 
This additional multimodal information offers valuable spatiotemporal and semantic cues, aiding individuals in locating objects of interest. 
Similarly, the ability to locate objects based on multimodal cues plays a crucial role in the functioning of machines.
For example, stage cameras should be able to locate the musician playing instruments, while movie cameras should locate the superhero in front of the yelling villain.
In the above scenarios, using a single modality for segmentation alone cannot achieve the desired behavior.

To date, researchers have studied the reference segmentation problem from somewhat limited scenarios. As shown in the coordinate system in Fig. \ref{fig:teaser}, there are currently three mainstream developments in segmentation methods for different modalities. 
Firstly, from the perspective of visual cues, Video Object Segmentation (VOS) \cite{ding2023mose,xu2018youtube,perazzi2016benchmark,perazzi2017learning,pont20172017} emerges with annotated first frame mask as the reference, guiding the segmentation of specific objects in subsequent video frames. While VOS can achieve arbitrary object segmentation in videos, it heavily relies on precise annotation of the first frame, which can be challenging and time-consuming in practice.
Secondly, from the perspective of text cues, Referring Video Object Segmentation (R-VOS) \cite{gavrilyuk2018actor,khoreva2019video,dutta2020unsupervised,Ding_2023_ICCV} explores the segmentation ability by using attribute descriptions as guidance. As R-VOS has successfully replaced the mask annotation used in VOS with natural language, it provides a more accessible and user-friendly reference form but the capability to locate objects in more natural dynamic audio-visual scenes is still limited. 
In contrast, from the perspective of audio cues, Audio-Visual Segmentation (AVS) \cite{zhou2022avs,zhou2023avss,yuanhong2023closer,liu2023annotation,wang2023prompting,zhou2023exploiting} leverages audio as a guide to segment sounding objects. This approach effectively addresses the challenge of locating objects in dynamic audio-visual scenes. However, AVS still faces limitations in its ability to focus on general objects that do not generate sound and cannot effectively locate objects of specific interest. 

In simple terms, current work falls short of enabling machines to locate objects of interest in natural dynamic audio-visual scenes.
For instance, as illustrated in \cref{fig:teaser}, how can machines accurately locate a person who is genuinely playing a musical instrument over time? This requires machines to infer which instrument is making a sound and who is playing that instrument. 
Though Yan \etal \cite{yan2023referred} utilize either audio or language modality for multimodal reference segmentation, their method does not integrate multimodal information simultaneously nor support the aforementioned scenario.
Therefore, introducing a task to explore the possibility of locating interested objects in natural dynamic audio-visual scenes holds significant practical potential.

In this paper, we propose a pixel-level segmentation task called \textit{\textbf{Ref}erring \textbf{A}udio-\textbf{V}isual \textbf{S}egmentation} (Ref-AVS), which requires the network to densely predict whether each pixel corresponds to the given multimodal-cue expression, including dynamic audio-visual information. 
Top-left of \cref{fig:teaser} highlights the distinctions between Ref-AVS and previous tasks. 
The Ref-AVS task poses a greater challenge as it requires the network to accurately locate and segment objects in a more complex and stereoscopic modality space.
To this end, a computation model with comprehensive multimodal understanding capabilities is necessary. 
To foster research advancements in this field, we introduce the \textit{Ref-AVS Bench}, the pioneering benchmark that focuses on locating and segmenting objects of interest using referring multimodal-cue expressions.
Considering the complexity of real-world audio-visual scenes, we collect about 4,000 audible video clips from YouTube of which more than 60\% are multi-source sound scenarios. More than 20,000 reference expressions are collected and verified by experts to ensure accuracy and reliability, which adopt multimodal cues to describe objects in diverse and dynamic audio-visual scenes. Moreover, a special unseen test set is considered to evaluate the model's generalization ability in the growing demand for zero-shot scenarios. 




Our contributions can be summarized as follows: 
\begin{itemize}
    \item 
    We propose Ref-AVS as a challenging scene understanding task that segments objects of interest with multimodal-cue expressions and provide the corresponding Ref-AVS benchmark for performance training and validation.
    
    \item We design an end-to-end framework for Ref-AVS that efficiently processes the multimodal cues with a crossmodal transformer, serving as a feasible research framework for future development.
    \item We conduct extensive experiments to demonstrate the advantages of considering multimodal cues for visual segmentation, which also indicate the superiority of our approach in all subsets. 
\end{itemize}

\section{Related Work}

\subsection{Referring Video Object Segmentation (R-VOS)}
The main objective of R-VOS is to perform object segmentation in streaming videos based on given natural language expressions. Initially proposed in 2018, A2D-Sentence \cite{gavrilyuk2018actor} aims to segment actors in video content based on descriptions of their actions. Subsequently, Refer-DAVIS17 \cite{khoreva2019video} and JHMDB-Sentences \cite{dutta2020unsupervised} use extra language as references to replace previous VOS settings.  
To facilitate large-scale R-VOS training, Seo et al. \cite{seo2020urvos} develop the Refer-YouTube-VOS dataset upon the YouTube-VOS-2019 dataset \cite{xu2018youtube} and Henghui et al. \cite{Ding_2023_ICCV} build the MeViS for complex motion expression. 
Overall, the primary R-VOS methods have evolved from semi-supervised VOS and incorporate attention mechanisms to enhance modalities interaction. 
For example, MAttNet \cite{yu2018mattnet} decomposes the expression information into language- and word-level attention, while URVOS \cite{seo2020urvos} introduces crossmodal attention and memory attention. ReferFormer \cite{wu2022language} and MTTR \cite{botach2022end} utilize the crossmodal transformer to connect language expression with image regions.
These datasets typically offer expressions that encompass the action or appearance attributes of the target object. 
However, these expressions primarily focus on visual information, which fails to meet the increasing multimodal demands in today's diverse audio-visual scenes.

\subsection{Audio-Visual Segmentation (AVS)}
The multimodal segmentation in audio-visual scenes, which is of great concern to us, has received some attention in AVS \cite{zhou2022avs,wang2023prompting,gao2023avsegformer}. 
AVS aims to obtain finer pixel-level masks corresponding to sound-emitting objects in the visual space. 
In contrast, previous studies on audio-visual localization \cite{Tian_2018_ECCV,hu2020discriminative} use heat maps for coarse localization in an unsupervised learning manner. 
The existing works on AVS can be broadly categorized into fusion-based methods \cite{zhou2022avs,liu2023audioaware,ling2023hear,huang2023discovering,gao2023avsegformer,li2023catr} and prompt-based methods \cite{wang2023prompting,mo2023av,liu2023annotation,ma2024steppingstones}. The former primarily focuses on localizing sounding objects by fusing audio and visual features, while the latter emphasizes constructing effective audio prompts for the visual foundation model. However, AVS simply segments all sound-emitting objects in the visual space, without the flexibility to combine the multimodal cues to segment specific objects of interest.

\subsection{Language-aided Audio-Visual Scene Understanding} 
Currently, there is a limited number of public tasks offering datasets for audio-visual scene comprehension accompanied by language assistance.
Two notable examples of such datasets are Music-AVQA \cite{Li2022Learning} and AVQA \cite{yang2022avqa}, both encompassing audio, visual, and language information.
These datasets contain rich audio-visual components and focus on annotations for textual questions and answers, emphasizing temporal (\texttt{Before}/\texttt{After}), spatial cues (\texttt{Left}/\texttt{Right}), \etc.

Researchers~\cite{pstpnet2023li, chen2023question, jiang2023target} acknowledge the significance of the provided question, as it guides the feature extraction process for both audio and visual signals. Consequently, they have made efforts to identify pertinent segments by evaluating the semantic similarity between the question and temporal segments.
Furthermore, \cite{pstpnet2023li} has been dedicated to locating crucial areas by leveraging semantic similarity between questions and visual tokens.
Clearly, existing explorations have significantly advanced the research on language-aided audio-visual scene understanding.
Nevertheless, the works that rely on the aforementioned datasets are still unable to offer pixel-level scene understanding. 
This encourages us to begin exploring fine-grained segmentation of dynamic audio-visual scenes in the real world by constructing appropriate datasets.

\section{Refer-AVS Dataset}

\subsection{Object Category}
To ensure a diverse range of referred objects, we have carefully selected a wide variety of audible objects in 48 categories as well as a smaller selection of static, unsoundable objects in 3 categories. For the objects that can produce sound, we have chosen 20 categories of musical instruments, 8 of animals, 15 of machines, and 5 of humans. In the special case of humans, considering their diverse appearances, voices, and actions, we have employed the concept of morphology and classified them into 5 categories, based on age and gender.

\subsection{Video Selection}


During the process of video collection, we employed the techniques presented in \cite{chen2020vggsound,zhou2022avs} to ensure the alignment of audio and visual snippets with the intended semantics. All videos were sourced from \textit{YouTube} under the \textit{Creative Commons license}, and each video was trimmed to a duration of 10 seconds. Throughout the manual collection process, we deliberately avoided videos falling into several categories (detailed in the appendix): 1) Videos with a large number of instances of the same semantics; 2) Videos characterized by extensive editing and camera switching; 3) Non-realistic videos containing synthetic artifacts.

To raise the alignment with real-world distributions, we carefully select videos that contribute to the diversification of scenes within our dataset. Specifically, we focus on selecting videos that involve interactions between multiple objects, such as musical instruments, people, vehicles, \etc. The rich combination of categories indicates that our dataset is not limited to a narrow set of scenarios but rather encompasses a broad spectrum of real-life scenes where such objects are likely to naturally appear together. Refer to the supplementary materials for details.

In addition to diversity, we also carefully filter the videos to ensure that the dataset includes scenes with greater complexity and a larger number of objects. Specifically, 56\% of the total videos contain two or more objects, while 13\% of the total videos contain three or more objects.

\subsection{Expressions}

\begin{figure*}[tb]
  \centering
   \includegraphics[width=1\linewidth]{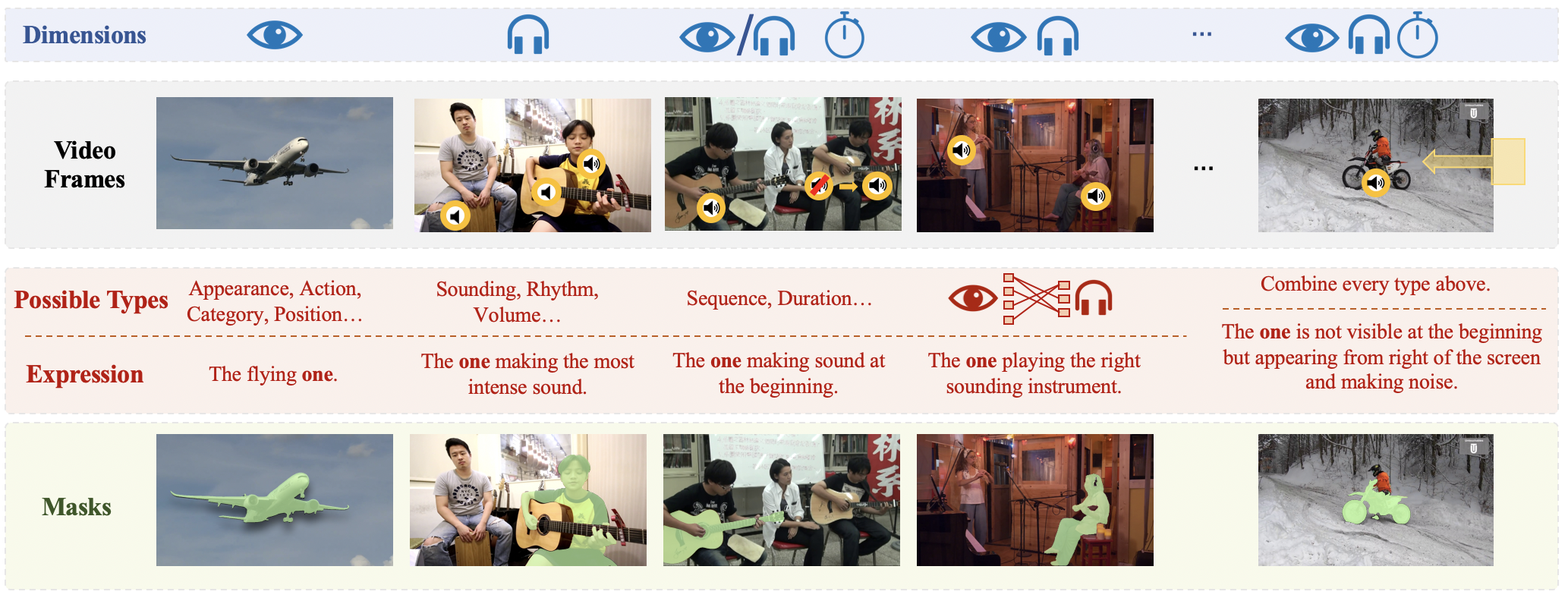}
    \caption{The illustration of our Ref-AVS benchmark. Note,for \includegraphics[height=0.3cm]{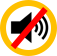} \includegraphics[height=0.3cm]{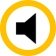} \includegraphics[height=0.3cm]{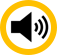}, the sound volume increases successively from silent to loud. Our benchmark is meticulously designed to encompass multimodal expressions from multiple dimensions. By combining various types of modality expressions, we achieve a dataset that exhibits great diversity.}
 
   \label{fig:dataset_type} 
\end{figure*}

The diversity of expressions is one of the core factors in the assembly of the Ref-AVS dataset. The expressions consist of three dimensions of information, namely audio, vision, and time. The auditory dimension incorporates characteristics such as volume and rhythm while the visual dimension encompasses attributes like the appearance and spatial configuration of objects. We also leverage temporal cues to generate references with sequential hints, \eg, \textit{``the one making sound first''} or \textit{``the one appearing later''}. By integrating auditory, visual, and temporal information, we craft a rich array of expressions that not only accurately reflect multimodal scenarios but also meet users' specific needs for precise references. 
\cref{fig:dataset_type} illustrates examples of the combination of different modalities.

The accuracy of expressions is also a core concern. We follow three rules to generate high-quality expressions: 
1) Uniqueness: The object referred to by an expression must be unique and an expression cannot refer to multiple objects simultaneously. 
2) Necessity: Complex expressions can be used for reference, but each adjective in the sentence should narrow down the scope of target objects, avoiding unnecessary and redundant descriptions of the object. 
3) Clarity: Certain expression templates involve subjective factors, such as \textit{``the \_\_ with a louder sound”}. The use of such expressions should only occur when the situation is clear enough to avoid ambiguous references.

Besides diversity and accuracy, we have graded the difficulty of expressions based on the number of cues involved, with the proportions of easy, medium, and hard samples in the overall dataset being 20\%, 60\%, and 20\%, respectively. This gradation could be a great benefit to future research like curriculum learning. Please refer to the supplementary material for more details.

\subsection{Segmentation Masks}
 
We divide each 10-second video into ten equal 1-second snippets and the objective of annotation is to acquire the first frame's mask for each snippet. For these sampled frames, the ground truth labels are binary masks indicating the target object, according to expressions and multimodal information.  

To obtain such a mask, initially, we manually select the pivotal frames for each 10-second video in which the target object is present. These pivotal frames may occur at the beginning, middle, or end of the video, depending on when the target object is most clearly visible. Subsequently, we utilize Grounding SAM \cite{ren2024grounded} to segment and label the pivotal frames, which will be manually checked and corrected subsequently. This process allows us to generate masks and labels for multiple target objects within the pivotal frames. Once the masks for the pivotal frames are established, we apply a tracking algorithm \cite{cheng2023tracking} to track the target object across the preceding and subsequent frames and obtain the ultimate mask and label for the target object in a span of 10 frames.

\subsection{Dataset Statistics}

\begin{table}[tb]
\centering 
\caption{Datasets comparison with related tasks.}
\scalebox{0.75}{
\begin{threeparttable}
\begin{tabular}{ccccccccc}
\toprule
Task                      & Dataset                    & Modality                   & Video & Frame   & Object                               & Expression               & Object/Video        & Pub.      \\ \hline
                          & Flickr-SoundNet~\cite{Senocak_2018_CVPR}            & A+V                        & 5,000  & 5,000    & -                                    & -                        & -                   & CVPR'2018 \\
\multirow{-2}{*}{AVL}     & VGG-SS \cite{chen2021localizing}                     & A+V                        & 5,158  & 5,158    & -                                    & -                        & -                   & CVPR'2021 \\ \hline
                          & {AVS~\cite{zhou2022avs}} & {A+V} & 12,356 & 82,972   & {13,500$ \sim $}             & {-} & 1.09$ \sim $                   & ECCV'2022 \\
\multirow{-2}{*}{AVS}     & {VPO~\cite{yuanhong2023closer}} & {A+V} & -     & 34,874   & {40,000$ \sim $} & {-} & 1.14 (Object/Frame) & CVPR'2024 \\ \hline
                          & J-HMDB Sentences~\cite{dutta2020unsupervised}           & T+V                        & 928   & 928     & 928                                  & 928                      & 1.28                & CVPR'2018 \\
                          & A2D Sentences ~\cite{gavrilyuk2018actor}              & T+V                        & 3,782 & 11,936  & 4,825                                & 6,656                    & 1                   & CVPR'2018 \\
                          & Refer-DAVIS$_{16}$~\cite{khoreva2019video}        & T+V                        & 50    & 3,455   & 50                                   & 100                      & 1                   & ACCV'2018 \\
                          & Refer-DAVIS$_{17}$~\cite{khoreva2019video}       & T+V                        & 90    & 13,543  & 205                                  & 1,544                    & 2.27                & ACCV'2018 \\
                          & Refer-Youtube-VOS \cite{seo2020urvos}         & T+V                        & 3,975 & 11,936  & 7,451                                & 27,899                   & 1.86                & ECCV'2020 \\
\multirow{-6}{*}{R-VOS} & MeViS~\cite{Ding_2023_ICCV}                     & T+V                        & 2,006 & 443,000 & 8,171                                & 28,570                   & 4.28                & ICCV'2023 \\ \hline
\textbf{Ref-AVS}                    & \textbf{Ref-AVS Bench}           & A+T+V                      & 4,002 & 40,020  & 6,888                                & 20,261                   & 1.72                & ECCV'2024      \\ 
\bottomrule
\end{tabular}
\end{threeparttable}
} 
\label{tab:datasets_comparison}
\end{table}

In addition, we compare Refer-AVS with other popular audio-visual benchmarks in the \cref{tab:datasets_comparison}. The Flickr-SoundNet \cite{Senocak_2018_CVPR} and VGG-SS \cite{chen2021localizing} benchmarks are labeled at a patch level through bounding boxes and possess around 5,000 frame-level annotations. Compared with our dataset with pixel-level annotations, these benchmarks have a significantly lower quantity of annotations. Regarding the AVS dataset \cite{zhou2022avs,zhou2023avss,yuanhong2023closer}, the videos in our dataset contain a higher average number of objects about 1.72 objects per video, implying that our scenarios are more challenging with multiple sound sources and multiple semantics. Within such scenarios, our Ref-AVS benchmark is particularly valuable as it demonstrates the ability to effectively focus on objects of real interest. Our dataset also has more uniform video durations and a more refined video selection process than AVS. When it comes to datasets for R-VOS tasks, we maintain a consistent advantage in the number of videos compared to other datasets \cite{gavrilyuk2018actor,khoreva2019video,dutta2020unsupervised,seo2020urvos,Ding_2023_ICCV}. Moreover, we possess a considerable volume of data encompassing a large number of objects, expressions, and complex scenes. 

Overall, our Ref-AVS dataset encompasses a substantial collection of 20,000 expressions and pixel-level annotations spread across 4,000 videos, totaling more than 11 hours. As a result, we believe that this benchmark adequately fulfills the requirements for facilitating research on the Ref-AVS task, while also presenting a significant level of challenge in this domain. We aim to continuously expand the dataset for broader community needs, like training larger foundation models. 

\begin{table}[tb]
\setlength{\tabcolsep}{6pt}
\centering
\caption{Dataset split of Ref-AVS Bench.}
\scalebox{1}{
\begin{threeparttable}
\begin{tabular}{c|c|ccc|ccc}
\toprule
\multirow{2}{*}{Subset} & \multirow{2}{*}{all} & \multirow{2}{*}{train} & \multirow{2}{*}{val} & \multirow{2}{*}{test} & test   & test     & test   \\
                        &                      &                        &                      &                       & (seen) & (unseen) & (null) \\ \hline
Category$\dagger$       & 52                   & 39                     & 39                   & 52                    & 39     & 13       & 1      \\
Video                   & 4,002                 & 2,908                   & 276                  & 818                   & 292    & 269      & 257    \\
Object                  & 6,888                 & 5,366                   & 518                  & 1,004                  & 478    & 269      & 257    \\
Expression              & 20,261                & 14,117                  & 1,349                 & 4,770                  & 2,288   & 1,454     & 1,028   \\ 
\bottomrule
\end{tabular}
\begin{tablenotes}
\item[$\dagger$] The categories here contains ``background''.
\end{tablenotes}
\end{threeparttable}
}
\label{tab:data_split} 
\end{table}

\subsection{Dataset Split}

As shown in \cref{tab:data_split}, the complete dataset is divided into three sets, \ie, a training set of 2908 videos, a validation set of 276 videos, and a test set of 818 videos. The videos in the test set and their corresponding annotations undergo a meticulous review and re-annotation process conducted by experienced workers. In order to thoroughly evaluate the models' performance in the Ref-AVS task, the test set is further divided into three distinct subsets, each serving a specific purpose.

\textbf{Seen}: The first test subset, referred to as the ``seen test set'', comprises object categories that have already appeared in the training set. The purpose of establishing this subset is to evaluate the model's fundamental performance and assess how well it generalizes to familiar object categories.

\textbf{Unseen}: To address the growing demand for the generalization capabilities of various models in an open-world scenario, an additional test set was created specifically to assess the model's ability to generalize to unseen audio-visual scenes. This ``unseen test set'' consists of object categories that did not appear in the training set, although their super-categories (\eg, animal, vehicle) may have been present in the training set. The purpose of this test set is to evaluate the model's capacity to generalize to novel object categories while leveraging its understanding of broader super-categories.

\textbf{Null}:  
A ``null reference problem'' arises when an expression refers to an object that either does not exist or is not visible in the given context. A model that accurately understands expression guidance should not segment any object in scenarios of a null reference \cite{wang2024segpref}. Based on this consideration, we have developed a ``null subset'' to test the model's robustness against such challenges. This subset comprises object categories that were present during training but are paired with expressions that do not correlate, ensuring all objects in the video frames are irrelevant to the given reference expression. Therefore, the true masks for this subset are empty, and the model should refrain from segmenting any object.  

\section{Expression Enhancing with Multimodal Cues}
\subsection{Overall Architecture}

Ref-AVS is designed to locate objects of interest in dynamic audio-visual scenes by utilizing multimodal cues. To solve this challenging task, we propose the Expression Enhancing with Multimodal Cues (EEMC) method, which focuses on integrating multimodal information in the dynamic audio-visual scenes into reference expressions with the corresponding multimodal cues. This approach enables the formation of a comprehensive multimodal reference feature. Additionally, we employ an attention mechanism to utilize the multimodal reference cues as a prompt for the visual foundation model, thereby facilitating the final segmentation process.

\begin{figure*}[tb]
  \centering 
   \includegraphics[width=1\linewidth]{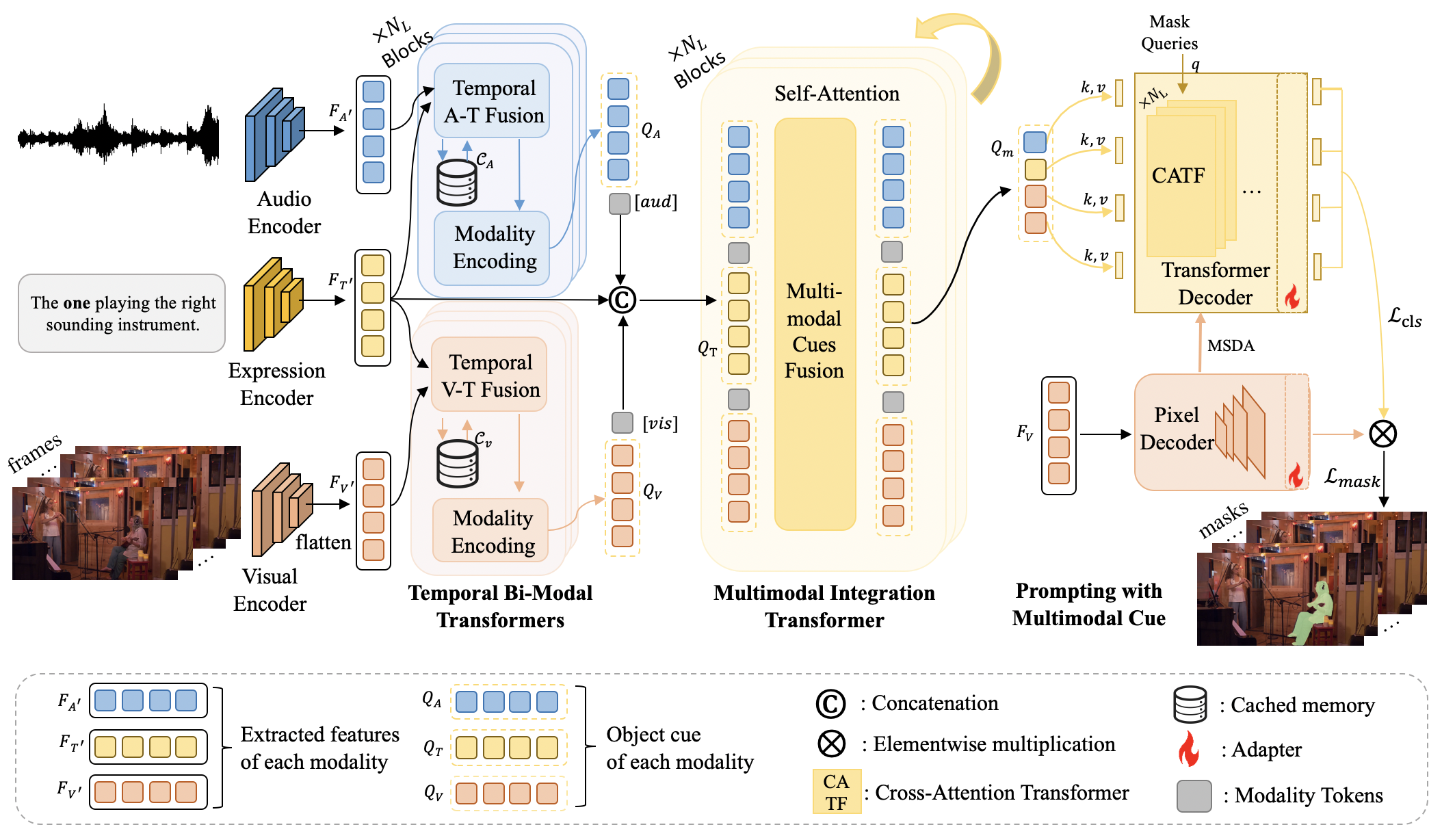}
   \caption{The overall architecture of our method EEMC.  We utilize cached memory to preserve temporal information, enabling the model to capture temporal mutations. Modality encoding provides a more clear context for the multimodal cue features $Q_m$, using the modality identify tokens $[aud]$ and $[vis]$. Lastly, we achieve efficient prompting of the visual foundation model by employing cross-attention, where mask queries $q$ serve as the target and multimodal cue features $Q_m$ act as the source.} 
   \label{fig:category}
\end{figure*}

\subsection{Multimodal Representations}
\textbf{Audio:} Similar to the video processing, we divide the audio input into clips at 1-second intervals. Audio representations $F_{A} \in \mathbb{R}^{t \times d_{A}}$ is encoded from VGGish \cite{gemmeke2017audio,hershey2017cnn}, where $t$ represents the duration of the audio in seconds and matches the number of video frames. The audio representations are extracted offline and the audio encoder is not fine-tuned.

\noindent\textbf{Visual:} We sample $t$ frames from the video input at 1-second intervals and extract visual representations $F_V \in \mathbb{R}^{t \times d_V \times H \times W}$ using a pre-trained Swin-base model \cite{dosovitskiy2020image}. The visual encoder is not fine-tuned.

\noindent\textbf{Expression:} We use RoBERTa \cite{devlin2018bert,liu2019roberta} as our text encoder to extract textual expression features $F_{T} \in \mathbb{R}^{L \times d_{T}}$, where $L$ denotes the length of the expression. The expression representations are obtained off-the-shelf without fine-tuning.

\subsection{Temporal Bi-modal Transformer}

\noindent \textbf{Temporal A-T \& V-T Fusion}: 
This module is introduced to retrieve the expression-related information of each modality. 
Firstly, for the convenience of the following multimodal fusion, we prepare the projected audio feature $F_{A^\prime} \in \mathbb{R}^{t \times 1 \times d_{V}}$, downsampled and flattened visual feature ${F_{V^\prime} \in \mathbb{R}^{t \times (\frac{W}{8}\times \frac{H}{8}) \times d_{V}}}$, and projected text feature $F_{T^\prime} \in \mathbb{R}^{t \times L \times d_{V}}$ which is expanded along the temporal dimension. Subsequently, the features are combined to yield the expression-related multimodal features $\hat{F}_{A}$, $\hat{F}_{V}$ and modality-aware expression features $\hat{F}_{T_{A}}$, $\hat{F}_{T_{V}}$, with the same dimension as ${F}_{A^\prime}$, ${F}_{V^\prime}$, ${F}_{T^\prime}$, respectively:

\begin{equation}
\hat{F}_{M},\hat{F}_{T_{M}}=MF(Concat(F_{M^\prime};F_{T^\prime})), M \in \{A, V\}
\end{equation}

\noindent where $Concat(\cdot)$ denotes the concatenate operation, and $MF(\cdot)$ denotes the modality fusion function, which is employed as self-attention.

\noindent \textbf{Cached Memory}: Although the aforementioned fusion method can combine cross-modal information, it does not give much emphasis to strong temporal dynamics. To address this issue, we introduce a straightforward cached memory $\mathcal{C}_A \in \mathbb{R}^{t \times 2 \times d_{T}}, \mathcal{C}_V \in \mathbb{R}^{t \times (W\times H+1) \times d_{T}}$ to explicitly capture the temporal mutations as they occur. The cache memory is required to store the mean modality features in the temporal domain from the beginning to the current moment. Then, the multimodal cues feature $Q_{A}$ and $Q_{V}$ output from temporal-aware transformers can be calculated as

\begin{align}
Q_{M}^\tau &= (\beta+1)\hat{F}_{M}^\tau - \beta\mathcal{C}_{M}^\tau, M \in \{A, V\}, \\
\mathcal{C}_{M}^\tau &=
\begin{cases}
 0, &\mbox{if } \tau=1,\\
 \frac{1}{\tau-1}\sum_{i=0}^{\tau-1} \hat{F}_{M}^i, &\mbox{if } \tau > 1,
 \end{cases}
\end{align}
while $\tau$ is the time stamp of the specific feature at the temporal dimension and $\beta$ is a hyper-parameter to control the sensitivity to temporal dynamics. When the current audio or visual features do not change significantly compared to the mean of past features (\ie, $\hat{F}_{M}^\tau=\mathcal{C}_{M}^\tau$), $Q_{M}^\tau$ remains nearly unchanged. 
However, in cases where the current change is drastic (\ie, $\hat{F}_{M}^\tau \neq \mathcal{C}_{M}^\tau$), the cached memory can amplify the difference in the current feature, leading to an output with salient features. It is important to note that text cue feature $Q_T$ inherently possesses a highly abstracted semantic nature, which means manipulation may not be necessary at this stage. 
Therefore, at this stage, we combine and average the original textual features with the text features enhanced by multimodal information, resulting in a comprehensive enhanced text feature $Q_T = mean(F_{T^\prime} + \hat{F}_{T_A} + \hat{F}_{T_V})$. Finally, we obtain the processed cue feature of each modality $Q_A$, $Q_V$ and $Q_T$.

\noindent \textbf{Modality Encoding}: The modality encoding approach involves incorporating modality identify tokens $[aud] \in \mathbb{R}^{1 \times d_V}$ for audio modality and $[vis] \in \mathbb{R}^{1 \times d_V}$ for visual modality into the following multimodal cues integration process. Since only two tokens are required to split the sequence into three segments, we only need to manipulate the audio and visual cue features.

\subsection{Prompting with Multimodal Cues}
In this Prompting with Multimodal Cues (PMC) phrase, when we obtain the final multimodal cues features, we can concatenate the multimodal cues together and then apply self-attention to obtain comprehensive multimodal cues $Q_m \in \mathbb{R}^{t \times (H \times W + L + 3) \times d_V}$ through the Multimodal Integration Transformer:
\begin{equation}
Q_m = MF(Concat([Q_A;[aud];Q_V;[vis];Q_T])),
\end{equation}
\noindent where $MF$ is the multimodal fusion function serving as the self-attention for multimodal cues interaction. Then we utilize these comprehensive multinational cues to prompt the learnable mask queries $q \in \mathbb{R}^{N \times d_V}$ in the transformer-based segmentation decoder with cross-attention, where $N$ is the number of mask queries:

\begin{equation}
q_Q = CATF(query=q, key=Q_m, value=Q_m),
\end{equation}
\noindent where $q_Q \in \mathbb{R}^{N \times d_V}$ is the updated mask queries. $CATF(\cdot)$ is the cross-attention transformer with $q$ and $Q_m$ as the information target and source, respectively. 

\section{Experiments}
\subsection{Implementation Details}
Mask2Former \cite{cheng2021mask2former} serves as our visual foundation model to provide the commonly used transformer-based segmentation decoder. 
By default, in this work, all input video frames are scaled to $384 \times 384$. The shape of the visual features is $[H=64, W=64, d_V=256]$, and we employ 8-fold downsampling to reduce the computation costs. The audio features with $d_A=128$ are extracted from the mono-channel waveform. The text features have a shape of $[L=25, d_T=768]$. For convenience, we map all modality feature dimensions to $d_V$. Hyper-parameter $\beta$ is set to 1 by default. Transformer layers $N_L$ of the Temporal Bi-modal Transformer, Multimodal Integration Transformer and the cross-attention transformer CATF are set to 4 by default. The number of mask queries $N_q$ is fixed to 100.
 
\subsection{Evaluation Metrics}

To conduct a comprehensive evaluation of our Ref-AVS method, we employ the Jaccard index ($\mathcal{J}$) and F-score ($\mathcal{F}$) as performance metrics. Additionally, we introduce $\mathcal{S}$ in the null reference test set to assess the efficacy of expression guidance in the model. $\mathcal{S}$ denotes the square root of the ratio between the predicted mask area and the background area, a higher value indicates a larger proportion of the mask relative to the background area, suggesting a lack of precise expression guidance provided to the model.

\subsection{Quantitative Results}
We compare our Ref-AVS method with existing approaches in the relevant field using our Ref-AVS benchmark. 
The results from the seen test set demonstrate the superior performance of our Ref-AVS method on this dataset, outperforming other methods by a significant margin. 
To ensure fairness, we equip the other methods with the same inputs from both the audio and visual modalities as ours. However, even with this additional information, these methods still fail to achieve satisfactory results. The result indicates that simple modal fusion is inadequate for addressing the challenges of understanding multimodal cues within the Ref-AVS task. 
Instead of directly incorporating audiovisual information, our approach selects textual representations as carriers of multimodal information because they contain rich semantics and cues that are contextually relevant to the current audiovisual environment.

We conduct tests on the unseen test set and the null reference test set to explore the generalization and multimodal cues-following ability. Our method still maintains a leading position in the unseen test set because we leverage text with highly abstract semantic capabilities as a multimodal information carrier, instead of directly fusing audio and visual information. Therefore, the obtained multimodal cues can provide more robust semantic guidance.
On the null test set, we obtain leading results among all other methods, indicating that our method can perceive multimodal cues to a reasonably accurate degree. Compared to other transformer-based and query-based frameworks, AVSBench, as a classic AVS method, obtains inferior results. This may be attributed to the model simply fuses multimodal cues and visual features in a progressive fashion, resulting in relatively weak guidance over the visual space.


\begin{table*}[tb]
\centering
 
\caption{Performance on Ref-AVS Bench. We use three test subsets to evaluate the comprehensive ability of Ref-AVS methods. \textbf{Mix} is the average performance of \textbf{Seen} and \textbf{Unseen} test set. We also use \textbf{Null} test set to evaluate the robustness of multimodal-cue expression guidance.}

\scalebox{0.95}{
\begin{tabular}{lccccccccccccc}
\toprule
\multicolumn{2}{c}{} & \multicolumn{2}{c}{Seen} & \multicolumn{2}{c}{Unseen} & \multicolumn{2}{c}{Mix(S+U)} & Null & Pub. \\ 


Method & Task & $\mathcal{J}$(\%) & $\mathcal{F}$ & $\mathcal{J}$(\%) & $\mathcal{F}$ & $\mathcal{J}$(\%) & $\mathcal{F}$ & $\mathcal{S}$ ($\downarrow$) & &  \\ 

\hline
AVSBench \cite{zhou2022avs} & \multirow{2}{*}{AVS} & 21.22 & 0.457 & 27.67 & 0.509 & 24.45 & 0.483 & - & \multirow{2}{*}{ECCV'2022} \\

\multicolumn{1}{c}{+\textit{text}} & & 23.20 & 0.511 & 32.36 & 0.547 &  27.78 & 0.529 & 0.208  \\
\hdashline[0.5pt/5pt]

AVSegFormer \cite{gao2023avsegformer} & \multirow{2}{*}{AVS} & 29.16 & 0.423 & 32.71 & 0.451 & 30.94 & 0.437 & - & \multirow{2}{*}{AAAI'2024} \\ 
\multicolumn{1}{c}{+\textit{text}} & & 33.47 & 0.470 & 36.05 & 0.501 & 34.76 & 0.486 & 0.171 &  &  &  \\
\hdashline[0.5pt/5pt]

GAVS \cite{wang2023prompting} & \multirow{2}{*}{AVS} & 24.85 & 0.455 & 27.86 & 0.460 & 26.36  & 0.457 & - &  \multirow{2}{*}{AAAI'2024} \\ 
\multicolumn{1}{c}{+\textit{text}} & & 28.93 & 0.498 & 29.82 & 0.497 & 29.38 & 0.498 & 0.190 &    \\
\hline

ReferFormer \cite{wu2022language} & \multirow{2}{*}{Ref-VOS} & 27.46 & 0.477 & 24.94 & 0.504 & 26.2 & 0.491 & - & \multirow{2}{*}{CVPR'2022} \\ 
\multicolumn{1}{c}{+\textit{audio}} & & 31.31 & 0.501 & 30.40 & 0.488 & 30.86 & 0.495 & 0.176 &  \\
\hdashline[0.5pt/5pt]

R2VOS \cite{li2023robust} & \multirow{2}{*}{Ref-VOS} & 21.34 & 0.309 & 23.91 & 0.431 & 22.63 & 0.370 & - & \multirow{2}{*}{ICCV'2023} \\  
\multicolumn{1}{c}{+\textit{audio}} & & 25.01 & 0.410 & 27.93 & 0.498 & 26.47 & 0.454 & 0.183 &  \\
\hline


EEMC & Ref-AVS & \textbf{34.20} & \textbf{0.513} & \textbf{49.54} & \textbf{0.648} & \textbf{41.87} & \textbf{0.581} & \textbf{0.007} & ECCV'2024 \\

\bottomrule
\end{tabular}
}
 
\label{tab:avs-v3}
\end{table*}




\subsection{Qualitative Results}
\begin{figure*}[tb]
  \centering
   \includegraphics[width=1\linewidth]{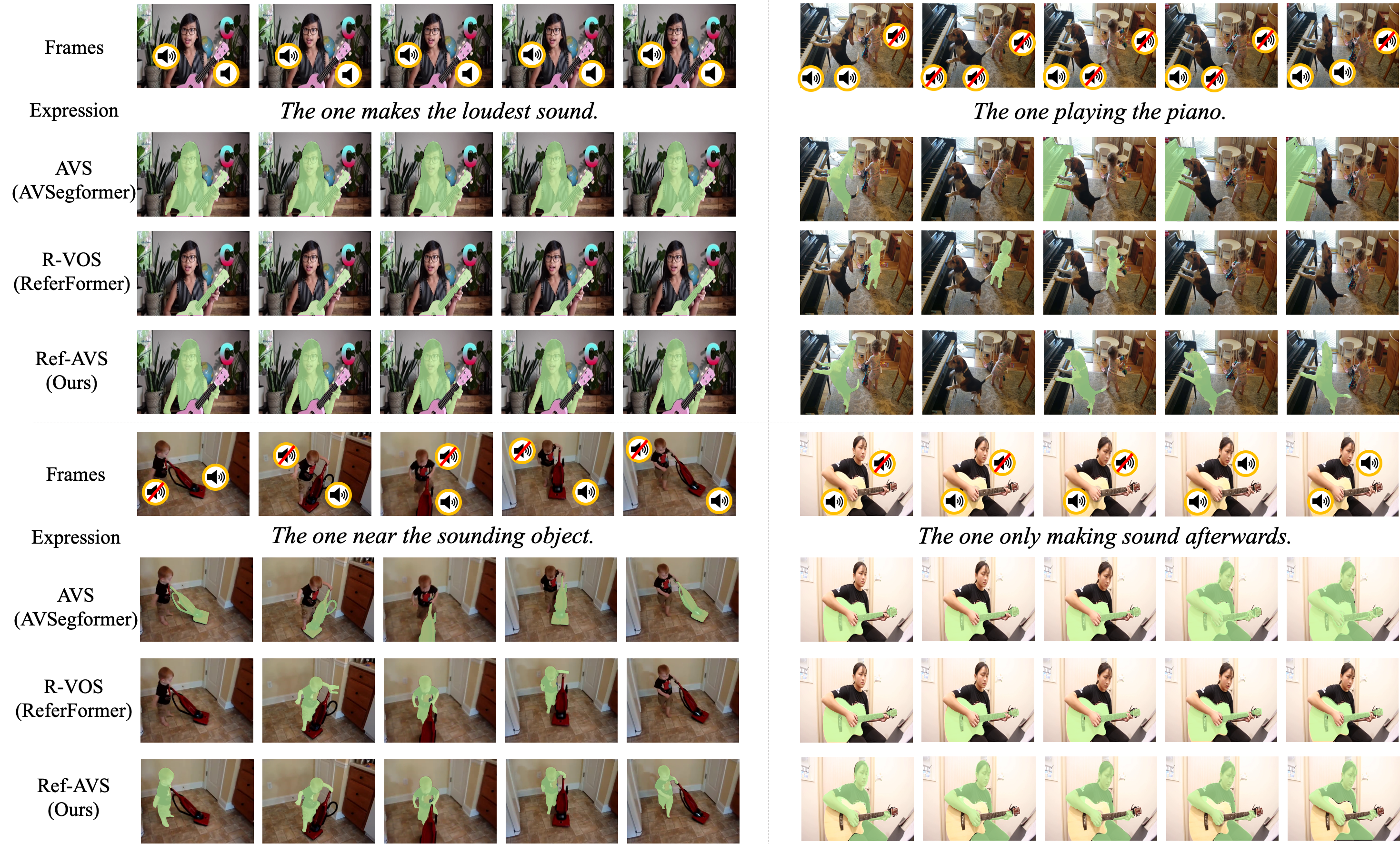}
   \caption{Segmentation masks of the referred objects in the Ref-AVS Bench. Note, for \includegraphics[height=0.3cm]{figs/emoji_mute.png}\includegraphics[height=0.3cm]{figs/emoji_low.png}\includegraphics[height=0.3cm]{figs/emoji_loud.png}, the sound volume increases successively from silent to loud. We compare our methodology with AVSegformer and ReferFormer. 
   More cases can be found in the supplementary material.}
   \label{fig:case}
 
\end{figure*}

We visualize the segmentation masks on the test set of the Ref-AVS Bench and compare them with AVSegFormer and ReferFormer, as shown in \cref{fig:case}. From these qualitative results, we observe that both ReferFormer in the Ref-VOS task and AVSegFormer in the AVS task fail to accurately segment the object described in the expression. Specifically, AVSegformer struggles to fully comprehend the expression and tends to directly generate the sound source. This issue is exemplified in the bottom-left sample, where AVSegformer erroneously segments the vacuum cleaner instead of the boy. On the other hand, Ref-VOS may not adequately understand the audio-visual scene and thus mistakenly identify the toddler as the piano player, as shown in the top-right sample. In contrast, our Ref-AVS method demonstrates a superior capability to simultaneously process multi-modal expressions and scenes, enabling it to accurately interpret the user instruction and segment the intended object of interest. 

\subsection{Ablation Study}
 

\begin{table*}[tb]
\setlength{\tabcolsep}{4pt}
\begin{center}
\caption{Ablation study on modalities and method modules.}
\label{tab:ablation}
\scalebox{0.9}{
\begin{threeparttable}
\begin{tabular}{lccccccc}
\toprule
\multirow{2}{*}{Setting} & \multicolumn{2}{c}{Seen} & \multicolumn{2}{c}{Unseen} & \multicolumn{2}{c}{Mix} & Null \\
 & $\mathcal{J}(\%)$ & $\mathcal{F}$ & $\mathcal{J}(\%)$ & $\mathcal{F}$ & $\mathcal{J}(\%)$ & $\mathcal{F}$ & $\mathcal{S(\downarrow)}$ \\ \hline
\ding{172} EEMC & \textbf{34.20} & \textbf{0.513} & \textbf{49.54} & 0.648 & \textbf{41.87} & \textbf{0.581} & \textbf{0.007} \\
\ding{173} (-) Text & 23.67 & 0.468 & 37.19 & 0.641 & 30.43 & 0.555 & 0.011 \\
\ding{174} (-) Audio & 27.27 & 0.492 & 43.64 & 0.656 & 35.46 & 0.574 & 0.010 \\
 \hline
\ding{175} (-) Cached Memory & 25.30 & 0.494 & 36.73 & \textbf{0.660} & 31.05 & 0.577 & 0.009 \\
\ding{176} (-) Modality Encoding & 27.20 & 0.470 & 44.97 & 0.637 & 36.09 & 0.554 & 0.009 \\ 
\bottomrule
\end{tabular}

 \end{threeparttable}
}
\end{center}

\end{table*}

We conduct ablation studies to investigate the impact of two modalities (audio and text) information on the Ref-AVS task, as well as the effectiveness of the proposed method.
As shown in the \cref{tab:ablation}, we observe that in setting \ding{173}, removing the text information results in a significant performance degradation of 11.44\% in $\mathcal{J}$ and 2.60\% in $\mathcal{F}$. This degradation is much higher compared to setting \ding{174}, where the removal of the audio information results in a combined mixed drop of 6.41\% in $\mathcal{J}$ and 0.70\% in $\mathcal{F}$. This phenomenon primarily results from the clarity and directness of textual information as a reference source. In contrast, when relying solely on audio information, the model often overlooks the referential content and instead focuses on identifying objects visually associated with audible behaviors.
We also conduct experiments to examine the function of each module in this framework. 
The cached memory is designed to track significant changes in the temporal domain. Meanwhile, the purpose of modality encoding is to extract more distinct and thorough features from multimodal cues, enhancing modal perception. Results from settings \ding{175} and \ding{176} in \cref{tab:ablation} demonstrate the effectiveness of these modules. 



%

\section{Conclusion}

In this work, we introduce a novel task, Ref-AVS, which requires the machine to segment objects of interest in dynamic audio-visual scenes based on expressions that incorporate multimodal cues and temporal information. 
To support this field, we develop the first benchmark, Ref-AVS Bench, for performance training and evaluation. 
Our method, EEMC, establishes a robust baseline with an advanced multimodal cues fusion module.
We compare our method with several existing approaches on the Ref-AVS Bench dataset and demonstrate its promising performance in accurately locating objects using multimodal reference expressions. 
Essentially, Ref-AVS leverages audio, visual, and textual data to enrich real-world scene comprehension. 
Our task has potential in applications such as attention focusing and object editing in immersive audio-visual scenarios like extended reality, thereby enhancing user interaction experiences.
In the future, we plan to further expand the scale of the current dataset to meet the growing data demands in the era of large language models.

\section*{Acknowledgements}
This research was supported by National Natural Science Foundation of China (NO.62106272), and Public Computing Cloud, Renmin University of China. This research was partially supported by Shanghai Artificial Intelligence Laboratory.

%
%
\bibliographystyle{splncs04}
\bibliography{main}

\newpage

\appendix

\title{Ref-AVS: Refer and Segment Objects in Audio-Visual Scenes \\
(Supplementary Material)}

\titlerunning{Ref-AVS}
\author{Yaoting Wang\inst{1}\inst{\dagger}\orcidlink{0009-0004-5724-5698}\thanks{$^\dagger$: Equal contribution.} \and
Peiwen Sun\inst{2}\inst{\dagger}\orcidlink{0009-0005-3016-8554} \and
Dongzhan Zhou\inst{3}\inst{\dagger}\orcidlink{0000-0001-6568-5440} \and
Guangyao Li\inst{1}\orcidlink{0000-0002-2179-8555} \and
Honggang Zhang\inst{2}\orcidlink{0000-0001-8287-6783}\and
Di Hu\textsuperscript{\Letter}\inst{1,4}\orcidlink{0000-0002-7118-6733} 
\thanks{\textsuperscript{\Letter}: Corresponding author.}
}

\authorrunning{Y.~Wang et al.}

\institute{
\textsuperscript{1} Gaoling School of Artificial Intelligence, Renmin University of China, China \\
\email{yaoting.wang@outlook.com} \\
\email{\{guangyaoli,dihu\}@ruc.edu.cn} \\
\textsuperscript{2} Beijing University of Posts and Telecommunications, Beijing, China \\
\email{\{sunpeiwen,zhhg\}@bupt.edu.cn}\\
\textsuperscript{3} Shanghai Artificial Intelligence Laboratory, Shanghai, China \\
\email{zhoudongzhan@pjlab.org.cn}\\
\textsuperscript{4} Engineering Research Center of Next-Generation Search and Recommendation\\
}

\maketitle

\section{Dataset}

\subsection{More statistics and examples}

The histogram in the diagram below represents the distribution of videos containing different categories of target objects. We can observe that the majority of the data is abundant and diverse. A small portion of the data is reserved as an unseen test set, whose categories have never appeared in the training set. Taking a broader view of the object categories, we can see that most objects are visually and acoustically related, with a few instances of silent and static data. We believe the diversity of object categories and quantity of the dataset will benefit future research in this challenging task. 

\begin{figure*}[!htbp]
  \centering
   \includegraphics[width=1\linewidth]{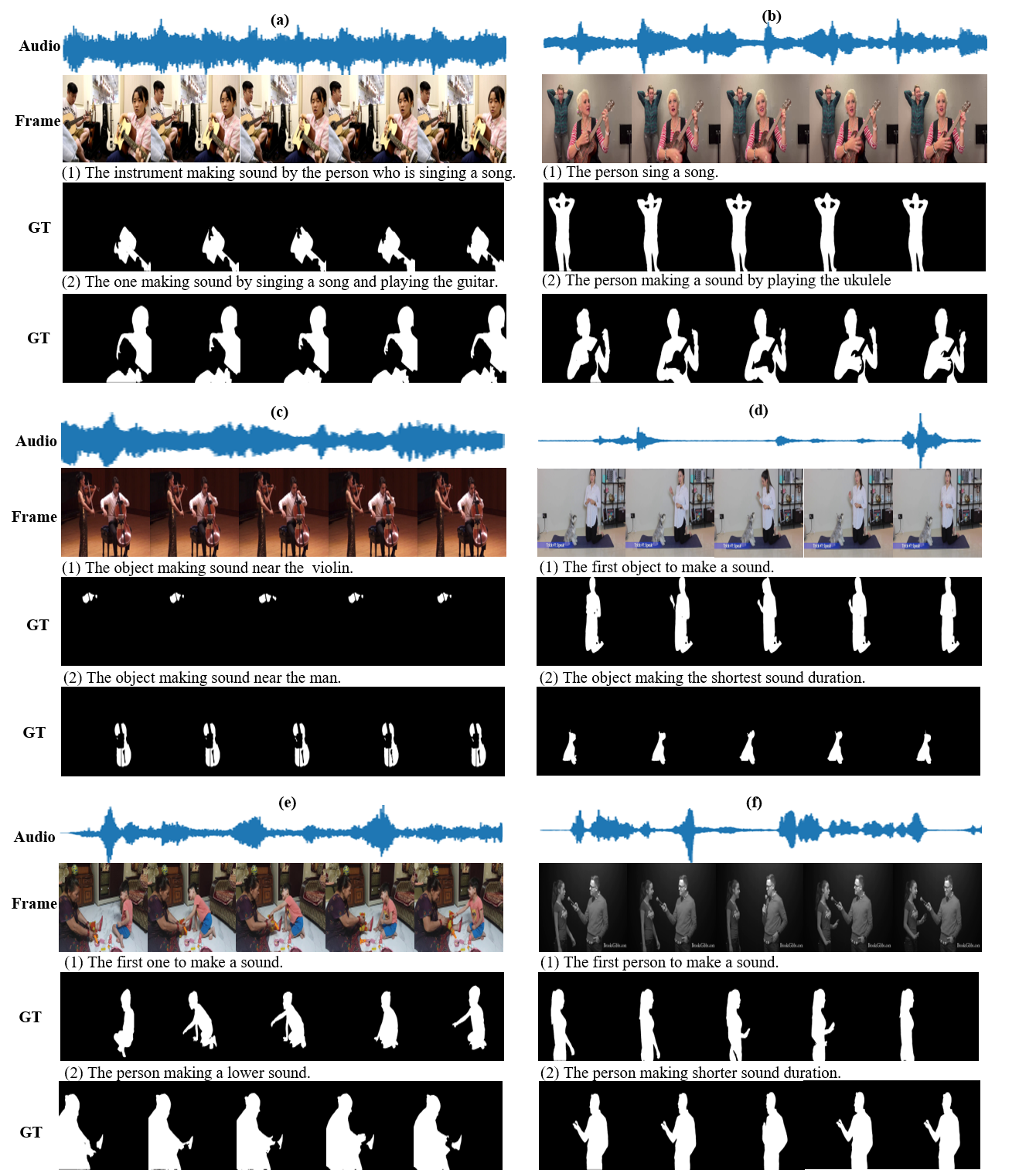}
   \caption{Data examples in our dataset Ref-AVS Bench. Use five time-steps for clarity.}
   \label{fig:samples}
\end{figure*}

As shown in \cref{fig:samples}, most videos in the dataset consist of challenging multi-source audio. Language reference expressions containing multimodal cues can assist us in locating objects of interest within audio-visual scenes.

\begin{figure*}[!htbp]
  \centering
   \includegraphics[width=1\linewidth]{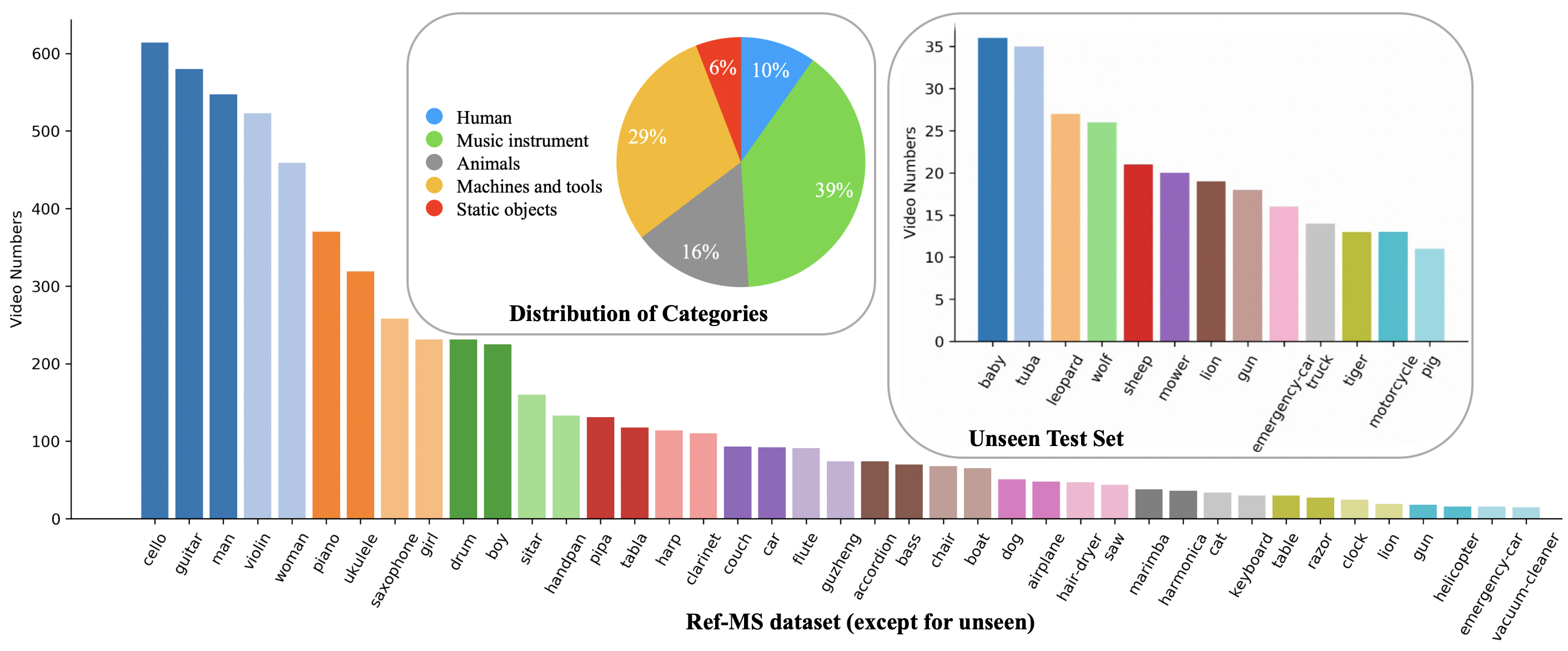}
   \caption{Illustration of the categories distribution for our Ref-AVS dataset, where seen and unseen categories are provided. The object classes were carefully chosen to ensure diversity and capture the distributional characteristics observed in real-world scenarios.}
   \label{fig:category}
\end{figure*}

\subsection{Data collection pipeline}

\begin{figure*}[tb]
  \centering
   \includegraphics[width=1\linewidth]{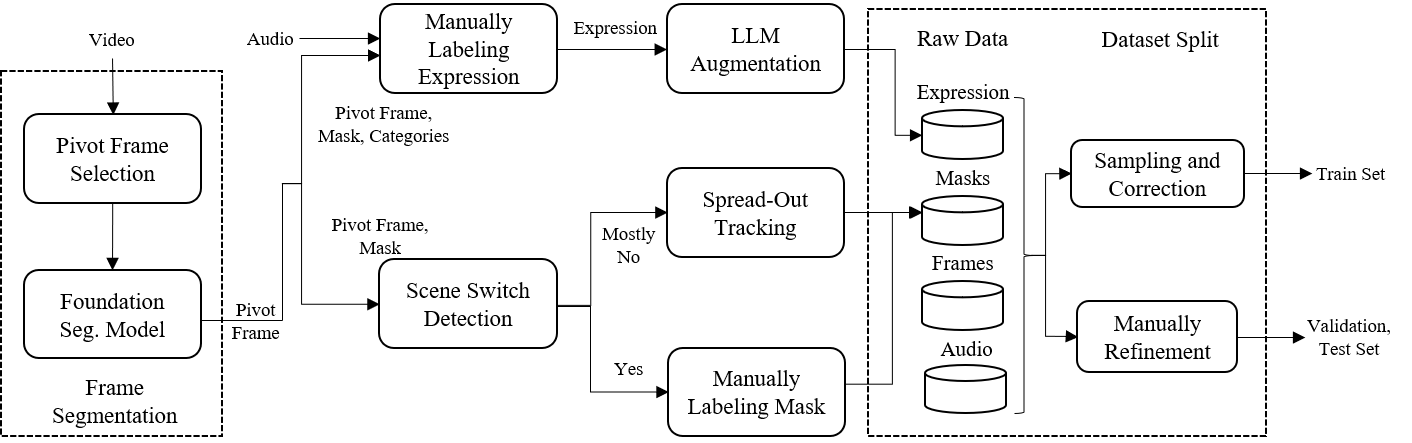}
   \caption{Dataset collection pipeline. The pipeline has played a significant role in ensuring the efficiency and cost-effectiveness of the overall process, leading to the successful acquisition of high-quality samples.}
   \label{fig:collection}
\end{figure*}

We carefully designed the rules to select the videos in the dataset, as mentioned in Sec.3.2 of the main text. Videos falling into one of the following categories will NOT be selected:
\begin{enumerate}
\item Videos with a large number of instances of the same semantics. Such videos often exhibit numerous overlapping instances and audio entanglement, making it challenging to accurately describe each instance, even for humans; 
\item Videos characterized by extensive editing and camera switching. These videos tend to showcase diverse scenery, causing constant changes in the relative positions of instances and the audio source. Consequently, the expressions are likely to be unclear and objects are more difficult to refer to. 
\item Non-realistic videos containing synthetic artifacts. Animation and hand-drawn videos involve secondary processing of object sound and images as part of their artistic expression.
\end{enumerate}
Following the rules above, we obtain the initial videos for further processing.
The overall data flow process is illustrated in \cref{fig:collection}. 

The process starts with the selection of a \textit{pivot frame} from the video, which is then used to detect and segment it using a \textit{foundation segmentation model}. Then, the \textit{scene switch detection} step is included to identify when the subject or context changes in the video. If a scene switch is detected, the process repeats, starting with the selection of a new pivot frame. This is followed by manual labeling for mask refinement and spread-out tracking for more accurate expression tracking.
Moreover, the expression is then augmented, which involves LLM to increase the diversity of the dataset and then split it into train and test sets. The train set is used to train a model, while the test set is used to evaluate the model's performance. Manual refinement is also applied to the validation set and test set.

The goal of the entire pipeline design process is to incorporate as many efficient frozen and off-the-shelf models as possible to minimize manual annotation costs while improving efficiency and accuracy. However, experienced annotators are still irreplaceable and are assigned to critical stages to ensure quality control and address specific issues such as tracking inefficiency caused by minor scene switches.
Finally, manual inspections on the validation (val) and test datasets are conducted by skilled researchers to ensure that they meet the community's requirements in terms of accuracy and quantity. For all videos in our dataset (train \& test), we require annotators to select quality masks from the auto-generated candidates. During test set verification, 156 (3.27\%) masks are manually corrected with less than 20\% pixel, indicating the quality of the mask annotation is acceptable to model training.

\subsection{More details for expressions}

\subsubsection{Templates.}
The initial expressions are generated via pre-defined templates, as shown in \cref{fig:template}. The templates are created by integrating messages from the audio, visual, and temporal dimensions. To avoid confusion, we provide the annotators with several examples so that they can fully understand the templates. 

We further define multiple placeholders that should be replaced in the templates, which can help the annotators produce suitable expressions more easily. The placeholders are summarized in \cref{tab:placeholder}. The usage of templates and placeholders enables flexible control and statistical analysis of the expressions, ensuring the quality of the dataset.  

In addition to the existing templates, we have granted experienced annotators and researchers access to a limited set of non-template expressions. This small collection of expressions brings higher flexibility beyond the constraints imposed by the used templates.

\subsubsection{Expression grade.}
\begin{figure*}[htbp]
  \centering
   \includegraphics[width=0.8\linewidth]{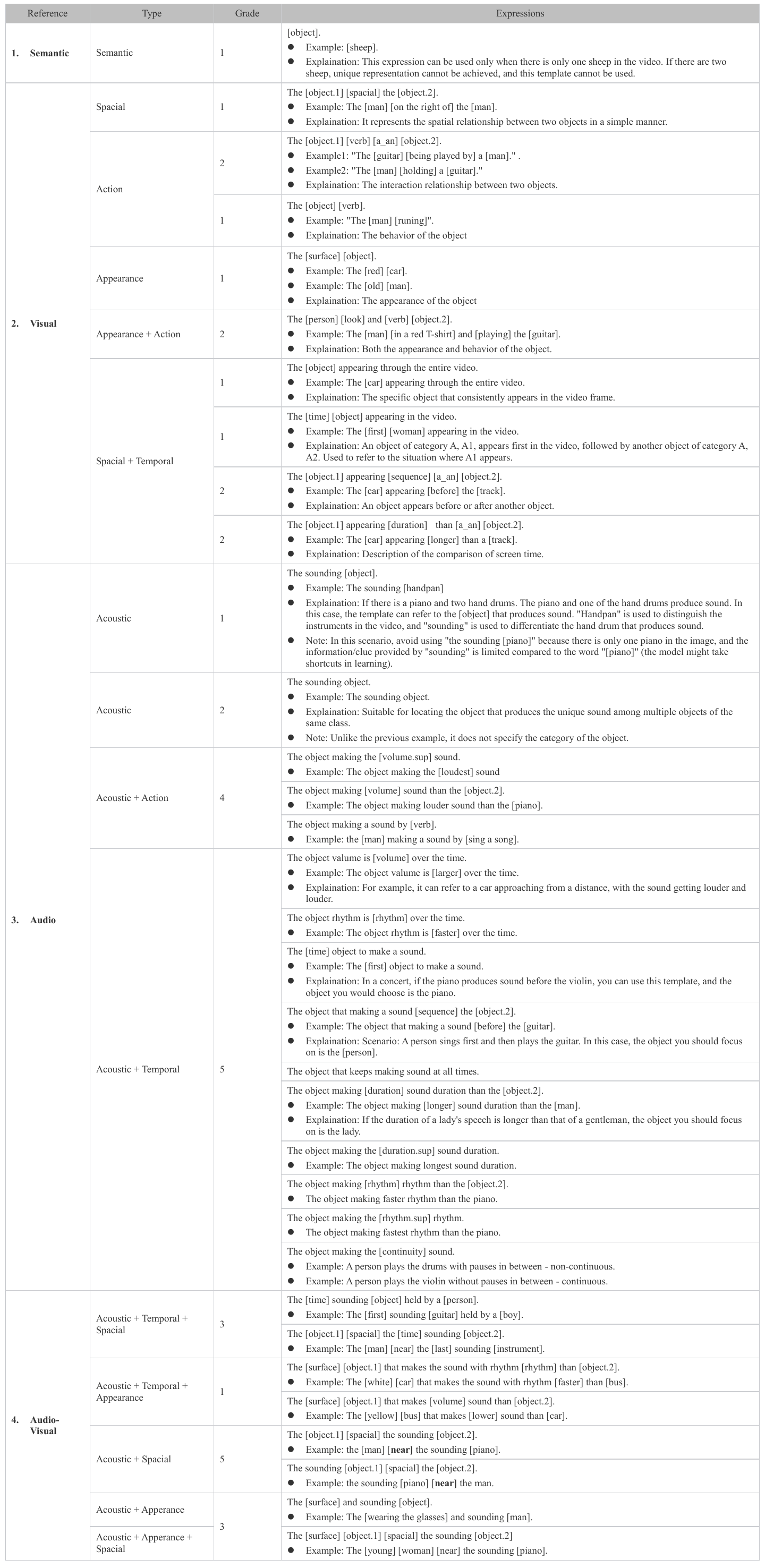}
   \caption{This chart shows the template we offered to the annotators. The third column of the chart shows a comprehensive evaluation of the grading system for expressions.}
   \label{fig:template}
\end{figure*}

We have implemented a practical grading system within the expression template, as shown in the third column of \ref{fig:template}. This grading system serves a dual purpose: it enables annotators and researchers to evaluate the difficulty of the data, while also establishing a strong foundation for future curriculum learning and related research endeavors.


\begin{table}[tb]
\centering
\caption{The list of placeholders. The annotators need to choose the proper words to replace the placeholder.}

\label{tab:placeholder}
\resizebox{0.6\textwidth}{!}{%
\begin{tabular}{llcc}

\toprule

Placeholder                   & Sub-label                & Semantic                                                  &  \\ \hline
\multirow{2}{*}{[temporal]}  
                              & [sequence]               & before, after;                                            &  \\
                              & [duration]               & longer, shorter;                                          &  \\ \hline
\multirow{2}{*}{[acoustic]}  
                              & [volume]                 & louder, lower;                                            &  \\
                              & [continuity]             & intermittent, continuous                                  &  \\ \hline
\multirow{5}{*}{[spacial]}    & [left-right]               & on the left of, on the right of;                          &  \\
                              & \multirow{2}{*}{[up-down]} & on, under;                                                &  \\
                              &                          & above, beneath;                                           &  \\
                              & [distance]                 & near, far;                                                &  \\
                              & [back-front]               & in front of, behind;                                      &  \\ \hline
\multirow{3}{*}{[appearance]} & [color]                  & red, green, ...;                                            &  \\
                              & [shape]                  & square, circle;                                           &  \\
                              & [look]                   & in a red t-shirt, wearing glasses, ...;    &  \\ \hline
[action]                      & [verb]                   & running, holding, being held by, ... &  \\ \hline
\multirow{2}{*}{[object]}     & [obj]                    & erhu, clock, piano, ...;                                       &  \\
                              & [person]                 & Specific category, \eg, man, boy, girl, ...;          &  \\
\bottomrule
\end{tabular}
}

\end{table}


\subsubsection{Uniqueness, necessity, and clarity.}

\begin{figure*}[tb]
  \centering
   \includegraphics[width=0.75\linewidth]{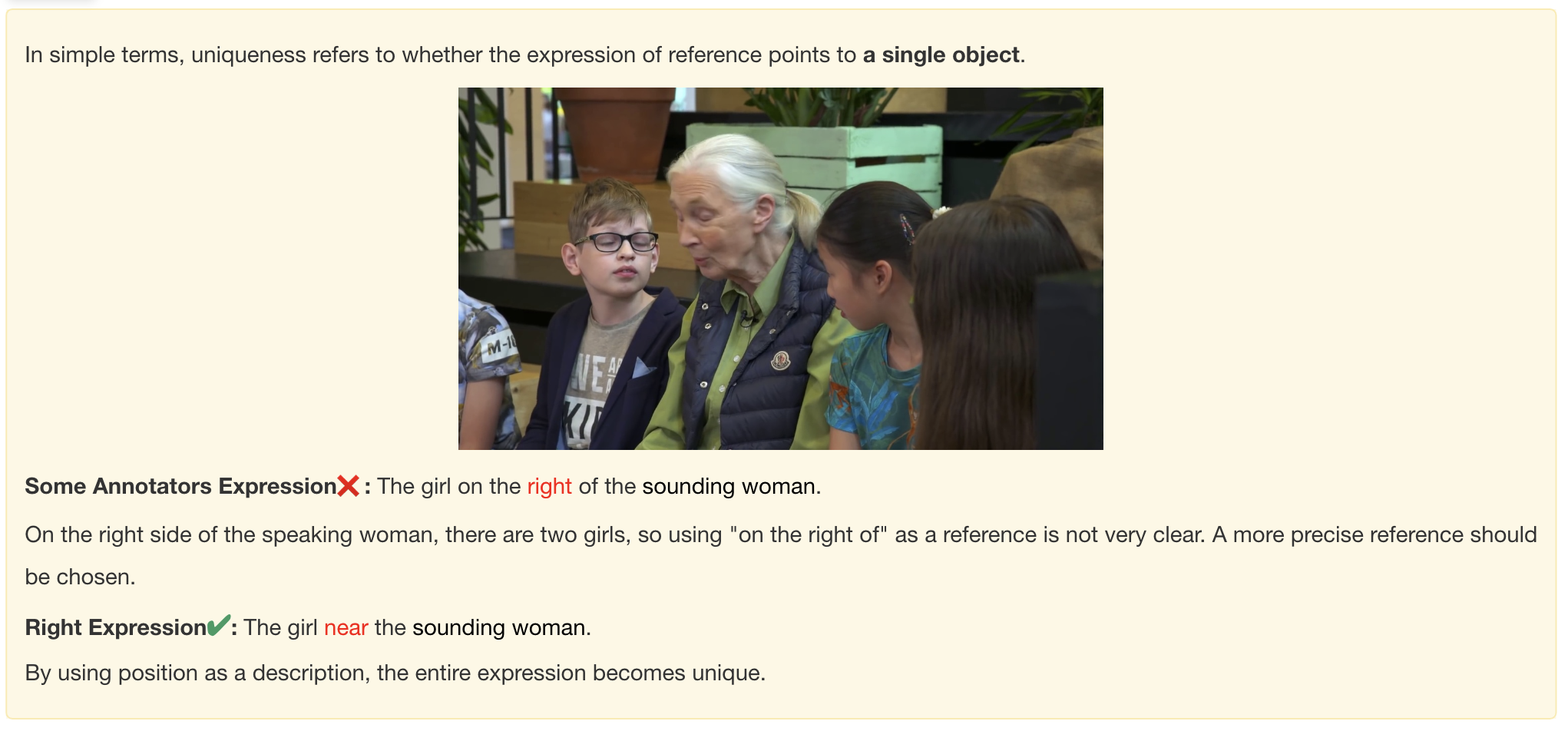}
   \caption{Uniqueness: The object referred to by an expression must be unique and an expression cannot refer to multiple objects simultaneously. }
   \label{fig:uniqueness}
\end{figure*}

\begin{figure*}[tb]
  \centering
   \includegraphics[width=0.75\linewidth]{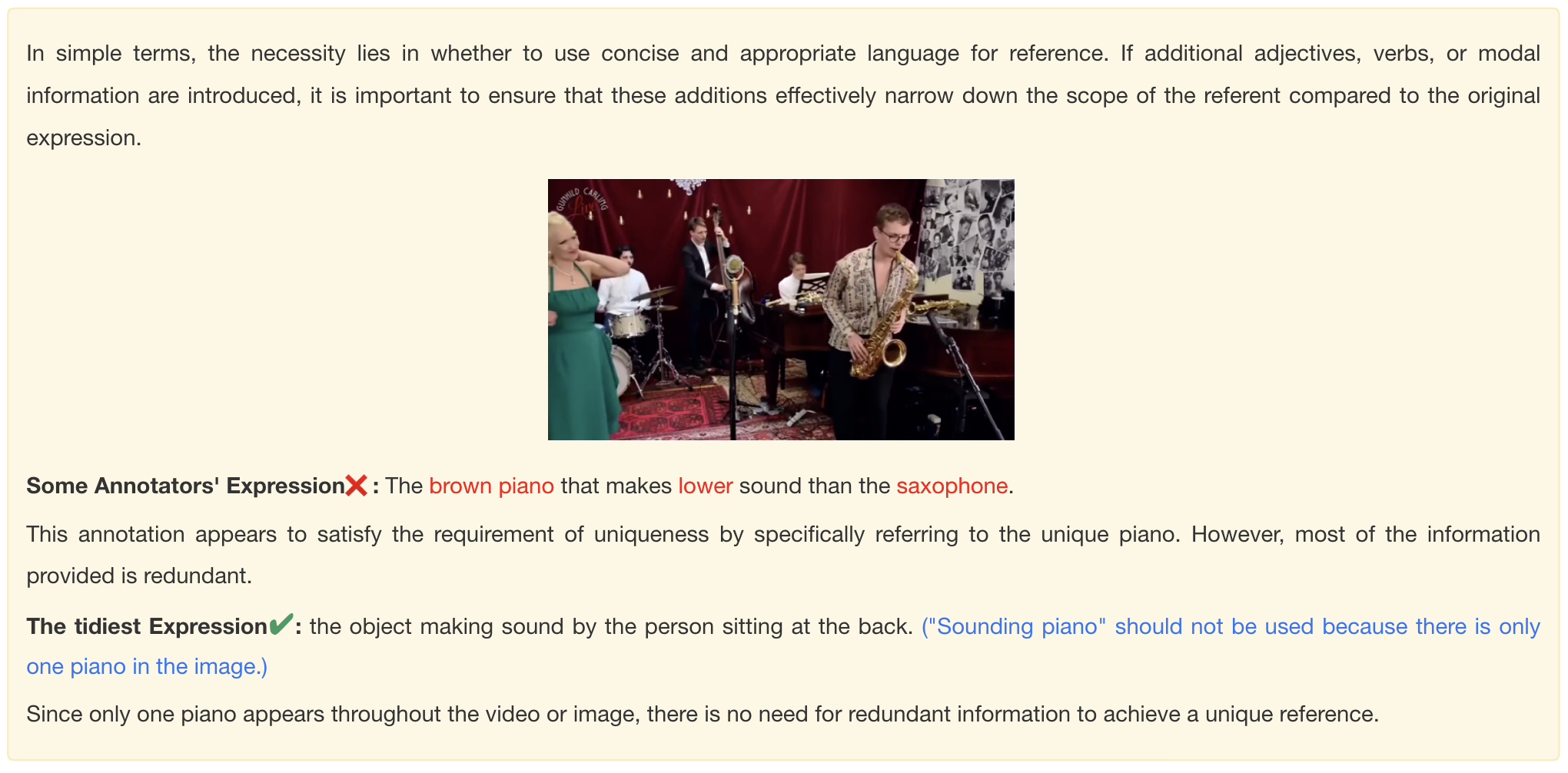}
   \caption{Necessity: Complex expressions can be used for reference, but each adjective in the sentence should narrow down the scope of target objects, avoiding unnecessary and redundant descriptions of the object. }
   \label{fig:necessity}
\end{figure*}

\begin{figure*}[tb]
  \centering
   \includegraphics[width=0.75\linewidth]{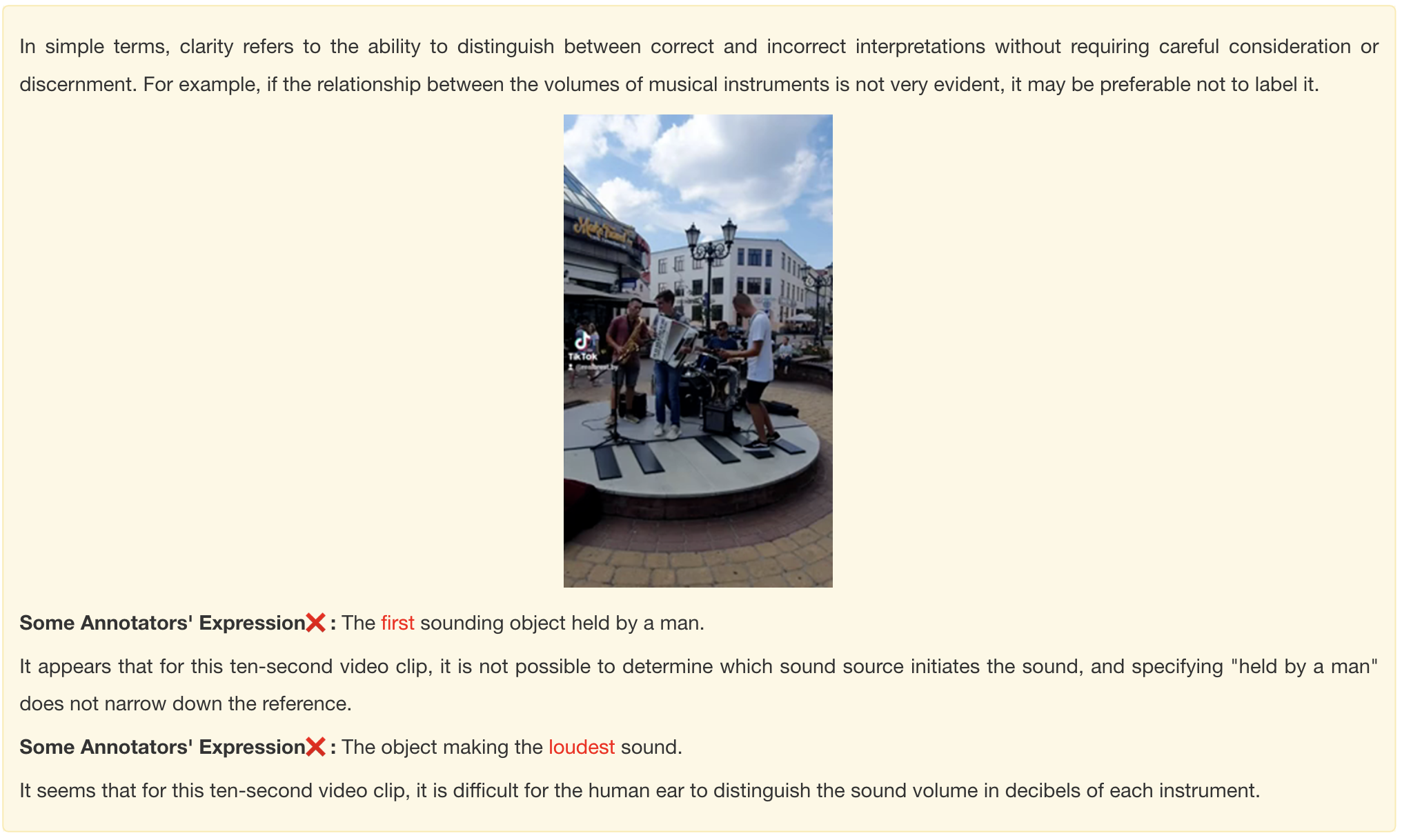}
   \caption{Clarity: Certain expression templates involve subjective factors, such as \textit{``the \_\_ with a louder sound”}. The use of such expressions should only occur when the situation is clear enough to avoid ambiguous references.}
   \label{fig:clarity}
\end{figure*}

As mentioned in Sec.3.3 in the main part, we provide three rules for expression annotation, namely uniqueness, necessity, and clarity. We introduce three cases to demonstrate these rules to the annotators, as shown in \cref{fig:uniqueness}, \cref{fig:necessity}, and \cref{fig:clarity}, respectively. 

\subsection{Dataset diversity}
\begin{figure}[tb]
  \centering
  \begin{subfigure}[b]{0.65\textwidth}
    \includegraphics[width=\linewidth]{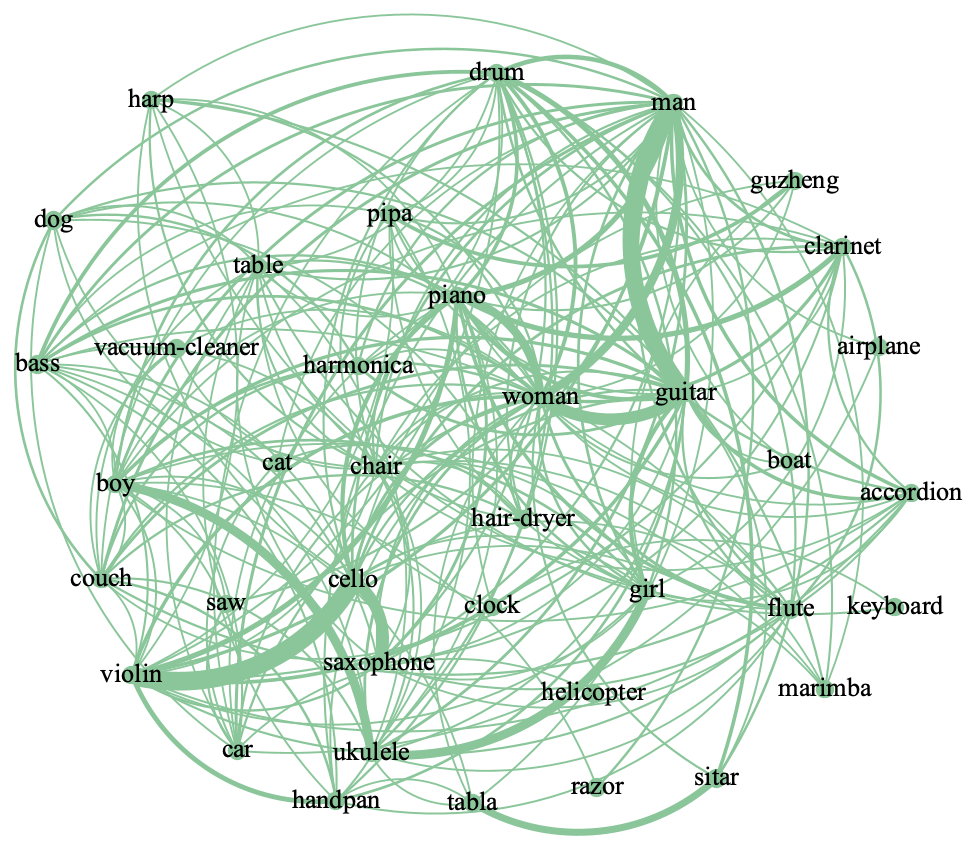}
    \caption{Object co-occurrence map}
    \label{fig:co-occurance1}
  \end{subfigure}
  \begin{subfigure}[b]{0.33\textwidth}
    \begin{subfigure}[b]{0.8\textwidth}
    \includegraphics[width=1\linewidth]{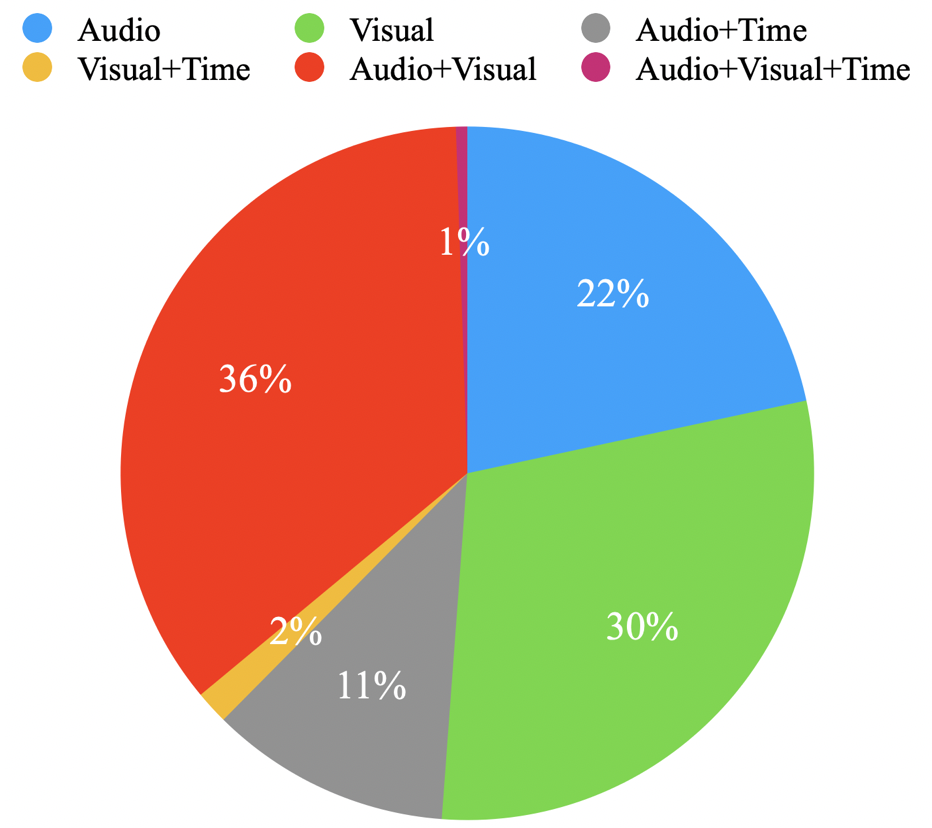}
    \caption{Multimodal cues distribution of expression}
    \label{fig:num_modality}
    \end{subfigure}
    \par\medskip
    \begin{subfigure}[b]{0.8\textwidth}
    \includegraphics[width=1\linewidth]{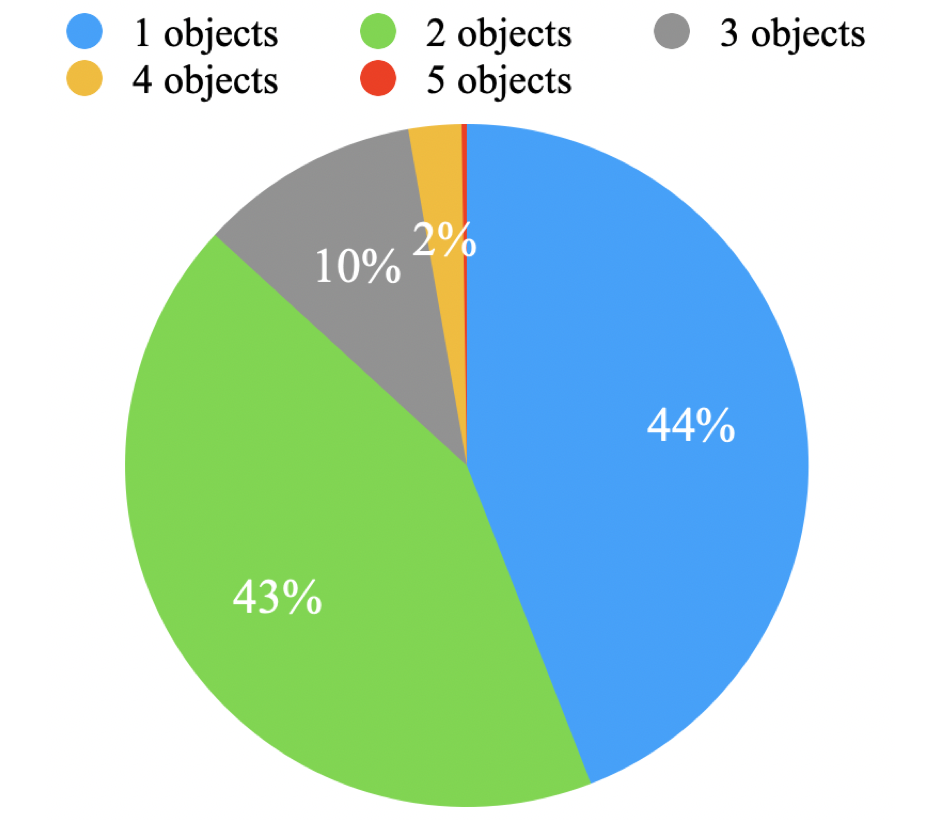}
    \caption{Objects over video distribution.}
    \label{fig:num_object}
    \end{subfigure}
    \label{fig:subfigure_layout}
  \end{subfigure}
  \caption{The distribution for both expressions and objects. We include multiple modalities combination and consider multiple objects appear in the same video.}
\end{figure}

\cref{fig:co-occurance1} visualizes the co-occurrence of objects in our dataset, where we can observe a dense web of connections spanning various categories, such as musical instruments, people, vehicles, \etc. The rich combination of categories indicates that our dataset is not limited to a narrow set of scenarios but rather encompasses a broad spectrum of real-life scenes where such objects are likely to naturally appear together. 
\cref{fig:num_object} demonstrates that 56\% of the total videos contain two or more objects, while 13\% of the total videos contain three or more objects.
The distribution of expressions in \cref{fig:num_modality} shows that expressions that combine two or more types of information account for nearly half. 

\subsection{Video duration}
The video length in our dataset is $10$s, which is a common setting in most audio-visual localization datasets (\eg, AVSBench, VGG-SS, and AVE) and is considered to contain ample audio-visual events. We collect videos from real-world scenes with high diversity to support our task, including various temporal contents (\eg, temporal order and event duration). 
\begin{figure}[!htbp]
    \centering
    \includegraphics[width=1\textwidth]{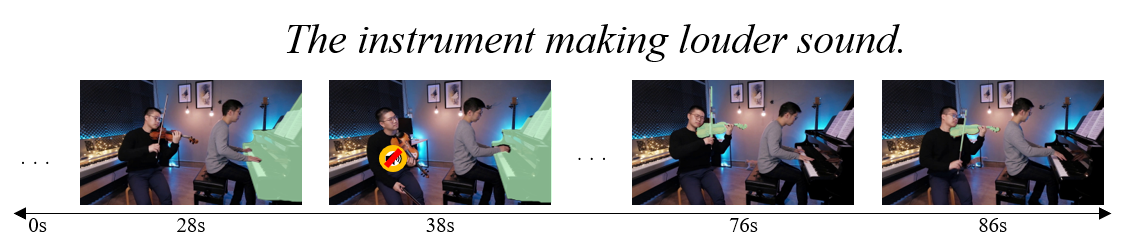}
     \caption{
     \small Ref-AVS on a long video around 90 seconds. In the video, the suonding object is changing along the temporal dimension.
     }
     \label{fig:long_range}
\end{figure}
As for the longer time interval setting, we test our model with a 90-second video, as shown in \cref{fig:long_range}. Despite being trained on 10s videos, our model exhibits generalization ability to videos of longer duration. Specifically, the output sticks to the object making a louder sound at the moment (piano first and then violin). However, building robust A-V correlations with long-range dependencies can still be challenging, which may be mitigated by grasping global audio-visual information and developing time-aware fusion methods. We leave the exploration for future work. 

\subsection{Dataset ablation} 
In \cref{tab:complexity}, we conduct further analysis for our dataset in terms of expression complexity and dimensions. We use the grade and length to reflect The complexity and difficulty of the expressions. Results in the first and second row of \cref{tab:complexity} show that mIoU decreases with harder expressions. As for the dimension, results in the last row of \cref{tab:complexity} illustrate that the acoustic subset is harder than the spatial subset. Moreover, the temporal dimension cannot exist alone, and adding temporal references may assist the segment process, indicating that time-aware modules like cached memory may be beneficial to this task.


\begin{table}[tb]
\centering
\caption{Dataset ablation and dimension analysis. Higher grade produce lower performance, same as the expression length.}
\resizebox{0.8\textwidth}{!}{
\begin{tabular}{ccccc}
\hline
Grade & 1+2 & 3 & 4 & 5 \\
mIoU & 39.27 & 37.78 & 35.31 & 33.39\\ \hline
Exp. len ($l$) & $l<=4$ & $4<l<=8$ & $8<l<=12$ & $ 12<l$\\
mIoU  & 39.66 & 37.00 & 33.74 & 29.61 \\ \hline
Dimension & \ding{172}spatial & \ding{173}acoustic & \ding{172}+temporal & \ding{173}+temporal\\
mIoU  & 33.83 & 26.07 & 41.01 & 32.30\\ \hline
\end{tabular}
}
\label{tab:complexity}
\end{table}

\section{Method}

\subsection{Learning objectives}
To enhance the quality of the mask during the model training process, we utilize both the binary cross-entropy loss and dice loss as optimization techniques:
\begin{equation}
\label{eq:loss_seg}
    \mathcal{L}_{mask} = \lambda_{bce} \cdot \mathcal{L}_{bce} + \lambda_{dice} \cdot \mathcal{L}_{dice}.
\end{equation}
Further, we adopt the mask classification loss to compose the final loss:
\begin{equation}
\label{eq:loss_seg}
    \mathcal{L}_{seg} = \mathcal{L}_{mask} + \lambda_{cls} \cdot \mathcal{L}_{cls}.
\end{equation}
The parameters $\lambda_{bce}, \lambda_{dice},\lambda_{cls}$ in the loss are set to $5$, $5$ and $2$ respectively, following the commonly used setting in mask transformer \cite{cheng2022masked} works.


\begin{table}[tb]
    \centering
    \caption{Ablation of different text encoder and audio encoder. (T) and (A) indicate the modified text and audio encoder.}
    \label{tab:encoder}
    \resizebox{0.8\textwidth}{!}{%
    \begin{tabular}{lcccc}
    \toprule
     & RoBERTa+VGGish & (T)Flan-T5 & (T)LanguageBind & (A)LanguageBind \\ \hline
    mIoU & \textbf{40.87} & 35.17 & 33.96 & 38.23 \\
    F-score & \textbf{0.581} & 0.547 & 0.540 & 0.561 \\
    \bottomrule
    \end{tabular}%
    }
\end{table}

\subsection{More ablation of our method}
In \cref{tab:encoder}, we conduct ablation experiments for text and audio encoders and find RoBERTa and VGGish work better. 

\subsection{Implementation detail on comparison method}
To incorporate missing audio or text modality information into other R-VOS or AVS methods, we employ linear projection to ensure that the newly added modality has the same feature dimensions as the existing modalities. Subsequently, we use an addition operation to combine the missing modality information with the original modality information. The remaining components of these methods remain unchanged in their default settings. To ensure fairness, all methods utilize the same audio encoder (VGGish \cite{gemmeke2017audio,hershey2017cnn}) and language encoder (RoBERTa \cite{devlin2018bert,liu2019roberta}).

\subsection{Supplementary video and more qualitative cases}

The attached file contains an \textit{MP4} video comprising seven segments. These segments encompass a variety of scenes (indoor, outdoor, concerts, \etc.), sound scenarios (single-source and multi-source), and temporal characteristics (temporally related and unrelated). We utilize our Ref-AVS Bench model to refer to and segment these videos, guided by the corresponding expressions. It is evident that our Ref-AVS model, Expression Enhancing with Multimodal Cues (EEMC), performs well in most cases. However, there are instances where the segmentation results are not as accurate, particularly in complex situations. Nevertheless, the video demo effectively showcases the challenges of the Ref-AVS task and demonstrates the effectiveness of our approach. 

\begin{figure}[tb]
    \centering
    \includegraphics[width=0.7\textwidth]{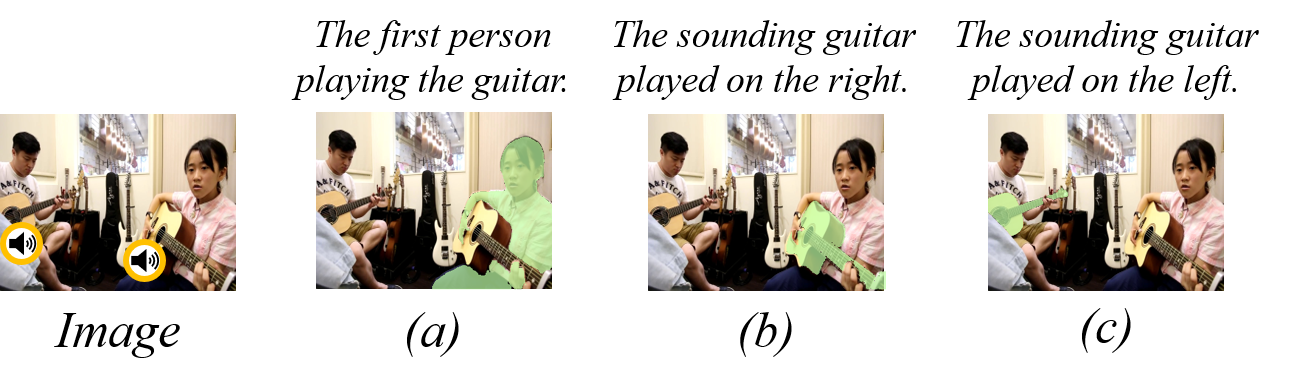}
     \caption{An example of the multi-instance scene. Text reference can help to disambiguate the real object of interested.
     }
     \label{fig:interaction}
\end{figure}

\begin{figure}[!htbp]
    \centering
    \includegraphics[width=1\textwidth]{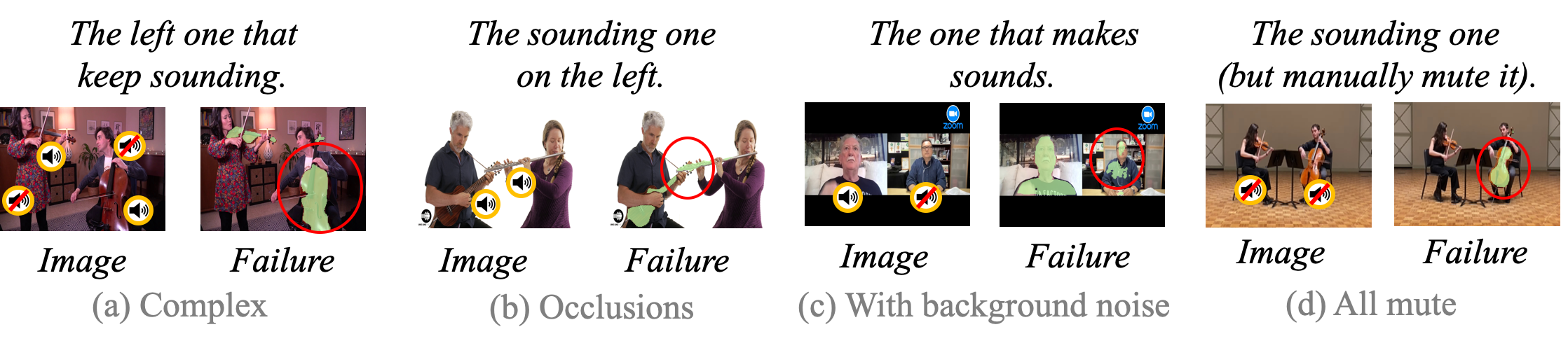}
     \caption{The failure cases for our model. Ref-AVS is a challenging task and many aspects is still in need of exploration.}
     \label{fig:bad_case}
\end{figure}

A variety of multi-instance scenes are covered in our dataset, including instruments, dialogue, \etc. As shown in \cref{fig:interaction}, our model successfully segments the target guitar. However, multi-instance scenes still pose challenges for our task. As shown in \cref{fig:bad_case}(c), the model generates a partial mask for the man who is not talking. The results illustrate that multi-instance scenes merit further research, especially in complicated scenes such as noisy backgrounds.

We provide more failure cases in \cref{fig:bad_case}(a), (b), (d) to demonstrate the challenge of our dataset. The model can be improved for complex scenery, occlusions, background noise, and wrong output in an all-mute scene.

\subsection{System latency}
The inference speed of our method is 2 FPS on 1$\times$RTX 3090. Pruning and quantization can be used to accelerate time-critical applications.


%
%

\end{document}